\documentclass[10pt,twocolumn,letterpaper]{article}

\usepackage{cvpr}              

\usepackage{graphicx}
\usepackage{amsmath}
\usepackage{amssymb}
\usepackage{booktabs}
\usepackage{bm}
\usepackage{multirow}

\usepackage[pagebackref,breaklinks,colorlinks]{hyperref}

\usepackage[capitalize]{cleveref}
\crefname{section}{Sec.}{Secs.}
\Crefname{section}{Section}{Sections}
\Crefname{table}{Table}{Tables}
\crefname{table}{Tab.}{Tabs.}

\def\cvprPaperID{*****} 
\def\confName{CVPR}
\def\confYear{2022}

\begin{document}

\title{Spatial Transformation for Image Composition via Correspondence Learning}

\author{$\textnormal{Bo Zhang}^{*}$,
$\textnormal{Yue Liu}^{*}$,
$\textnormal{Kaixin Lu}^{\dag}$,
$\textnormal{Li Niu}^{*}$, 
$\textnormal{Liqing Zhang}^{*}$\\
$^*$ Shanghai Jiao Tong University\,\,
$\dag$ Shanghai University \\
}
\maketitle

\begin{abstract}
   When using cut-and-paste to acquire a composite image, the geometry inconsistency between foreground and background may severely harm its fidelity. To address the geometry inconsistency in composite images, several existing works learned to warp the foreground object for geometric correction. However, the absence of annotated dataset results in unsatisfactory performance and unreliable evaluation. In this work, we contribute a Spatial TRAnsformation for virtual Try-on (STRAT) dataset covering three typical application scenarios. Moreover, previous works simply concatenate foreground and background as input without considering their mutual correspondence. Instead, we propose a novel correspondence learning network (CorrelNet) to model the correspondence between foreground and background using cross-attention maps, based on which we can predict the target coordinate that each source coordinate of foreground should be mapped to on the background. Then, the warping parameters of foreground object can be derived from pairs of source and target coordinates. Additionally, we learn a filtering mask to eliminate noisy pairs of coordinates to estimate more accurate warping parameters. Extensive experiments on our STRAT dataset demonstrate that our proposed CorrelNet performs more favorably against previous methods.
\end{abstract}

\section{Introduction}

\begin{figure}[t]
    \centering
    \includegraphics[width=0.9\linewidth]{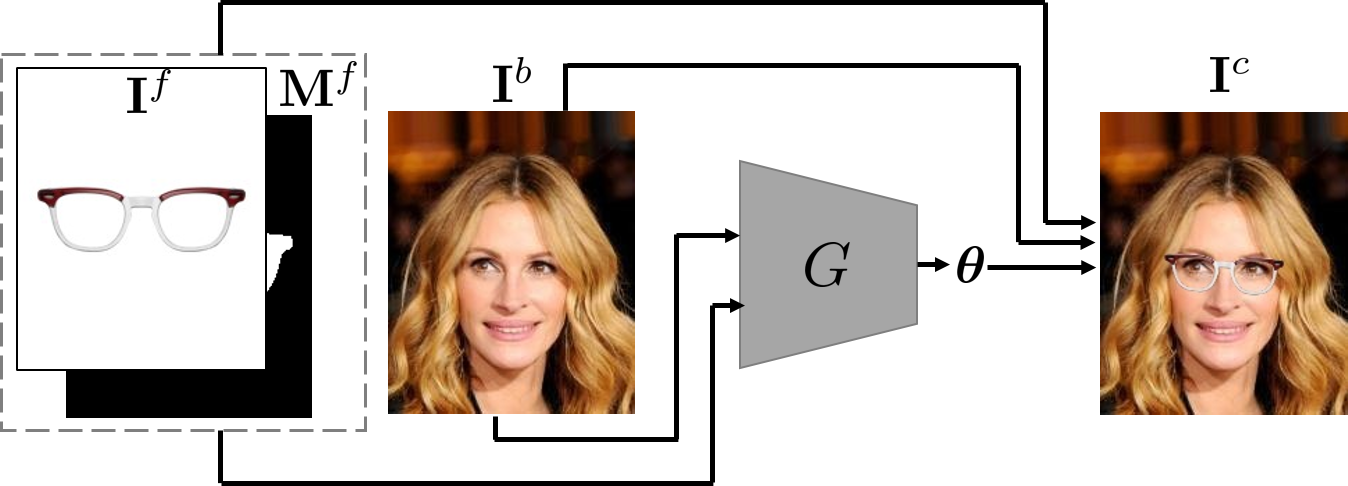}
    \caption{Illustration of spatial transformation for image composition. Given a pair of foreground $\mathbf{I}^f$ with mask $\mathbf{M}^f$ and background $\mathbf{I}^b$, the generator $G$ predicts the warping parameters $\bm{\theta}$ for the foreground. The warped foreground is combined with the background to produce a composite image $\mathbf{I}^c$.}\label{fig:task}
\end{figure}

\begin{figure*}[t]
    \centering
    \includegraphics[width=0.95\linewidth]{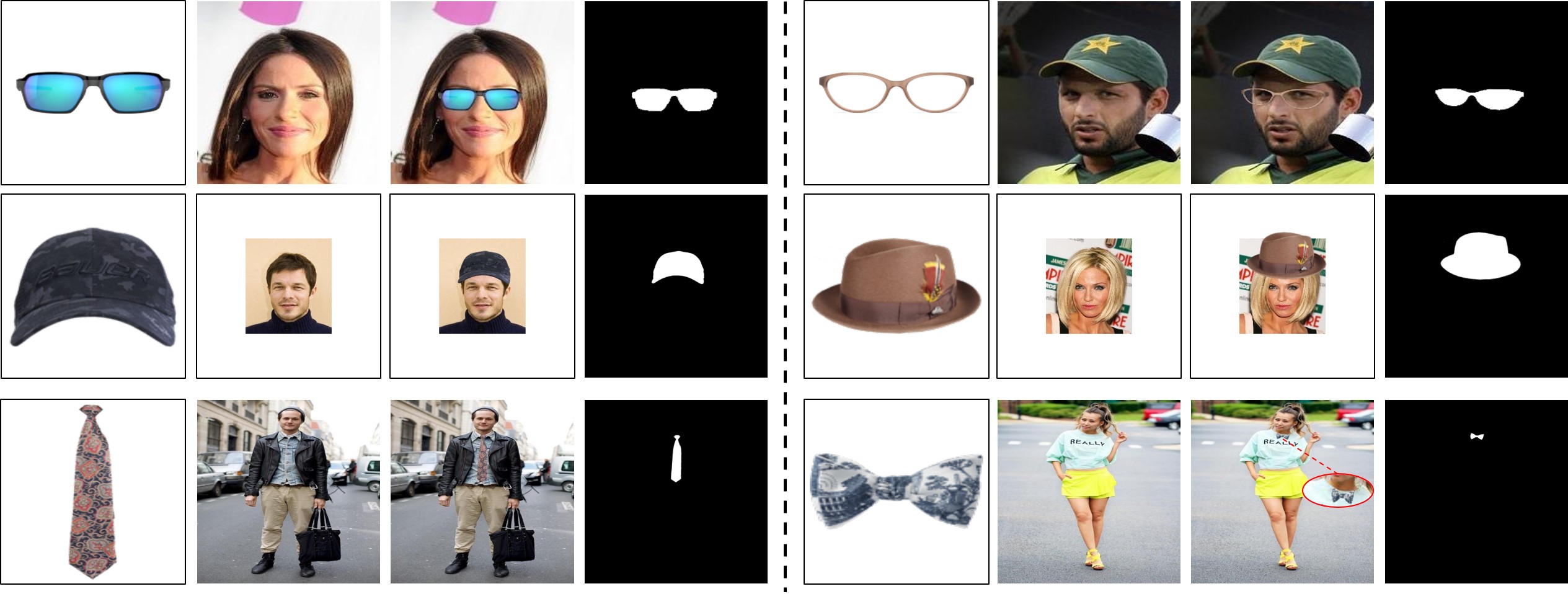}
    \caption{Examples in our STRAT dataset. From top to bottom: STRAT-glasses, STRAT-hat, STRAT-tie. We show two examples for each subdataset. In each example, from left to right: foreground image, background image, ground-truth composite image, ground-truth composite foreground mask. For STRAT-hat, we perform boundary padding for background image because the inserted hat may be out of the scope of original background. }\label{fig:dataset}
\end{figure*}

Image composition aims to cut the foreground from one image and paste it on another image to produce a composite image. As a common image editing operation, image composition has a wide range of applications such as virtual reality, artistic creation, automatic advertising~\cite{MISCWeng2020,WhatWhereZhang2020,niu2021making}. However, the obtained composite image may have inconsistent foreground and background, which would significantly harm the quality of composite image. The inconsistency between foreground and background can be divided into appearance (\emph{e.g.}, color, illumination) inconsistency and geometry (\emph{e.g.}, scale, location, camera perspective) inconsistency \cite{TowardRealisticChen2019,niu2021making}. Previous works on image composition target at solving one type or both types of inconsistency. 

In this work, we focus on solving the geometry inconsistency between foreground and background in a composite image. Previous works~\cite{stgan,RegGAN2019,SF-GAN,Compositional-GAN,GCC-GAN,Li2021ImageSV} on geometry inconsistency learned the warping parameter (\emph{e.g.}, perspective transformation) of foreground for geometric correction to create a realistic composite image (see Figure~\ref{fig:task}). Due to the lack of ground-truth annotations, these methods leverage weak supervision from real images, by using adversarial learning to enforce the produced composite images to be indistinguishable from real images. Despite the great progress they have achieved, their performances are still far from satisfactory. Moreover, the lack of reliable evaluation metrics also hinders the advance of this research field. Therefore, we contribute an annotated dataset with three subdatasets related to three virtual try-on applications, that is,  glasses try-on, hat try-on, and tie try-on. The reasons of choosing these three application scenarios are as follows. 1) Virtual try-on has broad application prospects and was studied in previous works \cite{stgan,RegGAN2019,Li2021ImageSV}. 2) The reasonable transformation has less uncertainty compared with other composition tasks (\emph{e.g.}, scene/text image synthesis \cite{Gaidon2016VirtualWorldsAP,Reed2016GenerativeAT}), which makes the ground-truth annotation more meaningful. 

As a pioneering dataset, we design a approach to collect the ground-truth annotation (see Figure~\ref{fig:annotation}). By taking the ``glasses try-on" subdataset as an example, given a pair of foreground (\emph{i.e.}, glasses) and background (\emph{i.e.}, human face), we ask human annotators to draw the foreground onto the background and adjust the four vertices of foreground to make its location, scale, and camera perspective reasonable. After adjustment, the coordinates of four vertices are recorded, based on which the ground-truth warping parameters of the foreground can be calculated. With such annotations, we can provide more informative supervision for the training process and design more reliable evaluation metrics. We name our dataset as Spatial TRAnsformation for virtual Try-on (STRAT) dataset. The details of dataset construction can be found in Section~\ref{sec:dataset}.

Previous works~\cite{stgan,RegGAN2019,SF-GAN,Compositional-GAN,GCC-GAN,Li2021ImageSV} on geometry inconsistency usually concatenate the foreground and background as input, which is delivered to the network to produce the warping parameters of the foreground. However, this type of network structure ignores the correspondence between foreground and background, which may provide useful insight for this task. For example, for ``glasses try-on", according to human experience, the middle of glasses should be roughly matched with the nose and the legs of glasses should be roughly matched with the ears. To capture such correspondence, we propose a novel correspondence learning network (CorrelNet) using foreground encoder and background encoder to extract the feature maps for foreground and background separately. Then, we calculate the cross-attention map between foreground and background feature maps, based on which we can predict the target coordinate that each source coordinate of foreground should be mapped to on the background. In light of pairs of source coordinates and target coordinates, we can directly calculate the warping parameters using linear regression. Considering that using all pairs of coordinates may involve redundant and noisy information, we also predict a filtering mask to discard partial pairs of coordinates for better warping parameter estimation. We compare our CorrelNet with previous methods on our constructed dataset, during which we also modify previous methods to better utilize the annotations in the dataset. The experiments show that our method outperforms the existing methods by a large margin. Our major contributions can be summarized as follows,
\begin{itemize}
    \item We contribute a Spatial TRAnsformation for virtual Try-on (STRAT) dataset, which consists of three subdatasets corresponding to ``glasses try-on", ``hat try-on", and ``tie try-on".
    \item We design a novel CorrelNet which captures the foreground-background correspondence and utilizes coordinate pairs to calculate the warping parameters. This technical route has never been explored before. We also develop a filtering strategy to further improve the performance. 
    \item Extensive experiments on our STRAT dataset demonstrate the superiority of our proposed method.
\end{itemize}

\section{Related Works}
In this section, we will firstly briefly review image composition methods which solve appearance or geometry inconsistency. Then, we will introduce some representative works on spatial transformation. Finally, we will mention the recent works on  virtual clothing try-on, which is related to our topic.

\subsection{Image Composition}
As a commonly used image editing operation, the goal of image composition is combining foreground and background as a realistic composite image. However, the fidelity and quality of composite images may be severely harmed by the inconsistency between foreground and background. According to \cite{TowardRealisticChen2019,niu2021making}, the inconsistency can be split into appearance inconsistency and geometry inconsistency. 

The appearance inconsistency means that the foreground and background have incompatible color and illumination statistics. To tackle this issue, image harmonization \cite{TsaiDIHarmonization2017,CongDoveNet2020,Ling2021RegionawareAI,Cong2021BargainNetBD,Zhu2015LearningAD} aims to adjust the color and illumination of foreground to make it compatible with the background, yielding a harmonious composite image. Besides, the inserted foreground may also have impact on the background such as casting shadow or reflection, so some works \cite{ARShadowGANLiu2020,hong2021shadow} attempted to generate plausible shadow for the inserted foreground. 

The geometry inconsistency means that the location, scale, or camera perspective of the inserted foreground are unreasonable. 
To tackle this issue, object placement methods~\cite{MVCIKAODDDvornik2018,WhereandWhoTan2018,WhatWhereZhang2020,DBLP:conf/eccv/ZhangWMWHS20,DBLP:conf/NeurIPS/LeeLG00K18} attempted to seek for the plausible location and scale for the inserted foreground. To further cope with inconsistent camera viewpoint, several spatial transformation methods for image composition~\cite{stgan,RegGAN2019,SF-GAN,Compositional-GAN,GCC-GAN} predicted the warping parameters of the foreground to enable more complicated transformation and make more fine-grained geometric correction. Analogous to \cite{stgan}, our work focuses on solving geometry inconsistency by predicting reasonable warping parameters for the inserted foreground object. 
To advance the research in this field, we construct an annotated dataset and propose a novel method which significantly outperforms previous methods.

\subsection{Spatial Transformation}
Spatial transformation is crucial in a variety of computer vision tasks (\emph{e.g.}, image classification, object detection, image generation) and has attracted considerable research interest. 
Spatial transformation can be roughly categorized into global spatial transformation and local spatial transformation. 
Global spatial transformation aims to learn transformation parameters for the whole image. For example, the representative spatial transformer network (STN) \cite{stns} learned the transformation parameters to convert image to prototypical shape. Inverse compositional spatial transformer network (IC-STN) \cite{Lin2017InverseCS} utilized the iterative strategy to achieve equivalently image alignment. Local spatial transformation aims to learn different transformation parameters for different regions and the transformation can be realized in various forms, which is more flexible than global spatial transformation. To name a few,  deformation convolution network (DCN) \cite{dai2017deformable,zhu2019deformable} predicted offsets for the convolution operation which can approximate the shape of target object. \cite{recasens2018learning} predicted distortion grid to emphasize the salient regions, which relocate pixels to the target coordinates. \cite{ren2020deep} used optical flow to spatially transform different regions for person image generation. 

Regarding the form of spatial transformation, our work belongs to global spatial transformation, which learns the warping parameters for the whole foreground. Regarding the application, we focus on a different application, that is, image composition. Therefore, our work is closer to previous image composition works  \cite{stgan,RegGAN2019,SF-GAN,Compositional-GAN,GCC-GAN} learning spatial transformation.  

\subsection{Virtual Try-on}
Virtual try-on techniques, which allow the customer to visualize the product on themselves before purchasing, can enhance shopping experience and improve customer satisfaction.
Existing virtual try-on techniques can be classified as 3D and 2D virtual try-on, both of which are important research fields.  As summarized in \cite{he2022fs_vton,Ge2021ParserFreeVT,feng2021weakly}, 3D virtual try-on provides better try-on experience (e.g., allowing being viewed with arbitrary views and poses), yet relies on 3D measurements or 3D parametric human body model, making it more challenging. In contrast, 2D approaches can be treated as more lightweight solutions. 
Most 2D virtual try-on works \cite{han2018viton,wang2018toward,zheng2019virtually,yang2020towards,ge2021parser,han2019clothflow,yang2021ct} focus on clothing try-on, that is, trying on the source clothes on the target person. With the assumption that the deformation between the clothing region and the corresponding region in the target person can be modelled by thin plate spine (TPS) transformation, many works \cite{han2018viton,wang2018toward,zheng2019virtually,yang2020towards} learned to predict TPS transformation to warp the source clothes. However, TPS cannot deal with large geometric changes and may unnaturally deform the source clothes. Thus, some recent works proposed more advanced transformation like dense warping \cite{yang2021ct} and appearance flow \cite{ge2021parser,han2019clothflow} to transform the source clothes.

Instead of clothing try-on in the above works, we focus on glasses, hat, and tie try-on, which we assume can be achieved by perspective transformation. Moreover, our task is an image composition task. This is also quite different from the above works which transfer the clothes on the target person to the source clothes. 

\begin{figure}[t]
    \centering
    \includegraphics[width=0.9\linewidth]{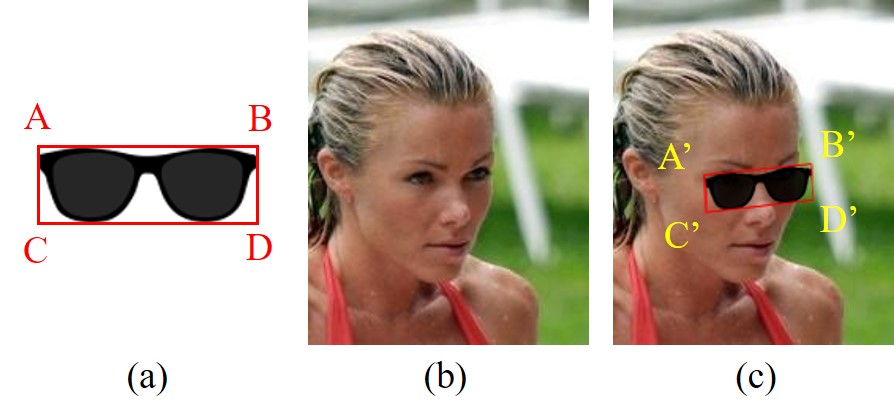}
    \caption{The illustration of annotation process when building our dataset. (a) The foreground object with four vertices A, B, C, D. (b) The background image. (c) The human annotator drags the foreground (a) onto background (b) and adjusts four vertices. After adjustment, four new vertices A', B', C', D' are recorded. }\label{fig:annotation}
\end{figure}

\begin{figure*}[t]
    \centering
    \includegraphics[width=0.8\linewidth]{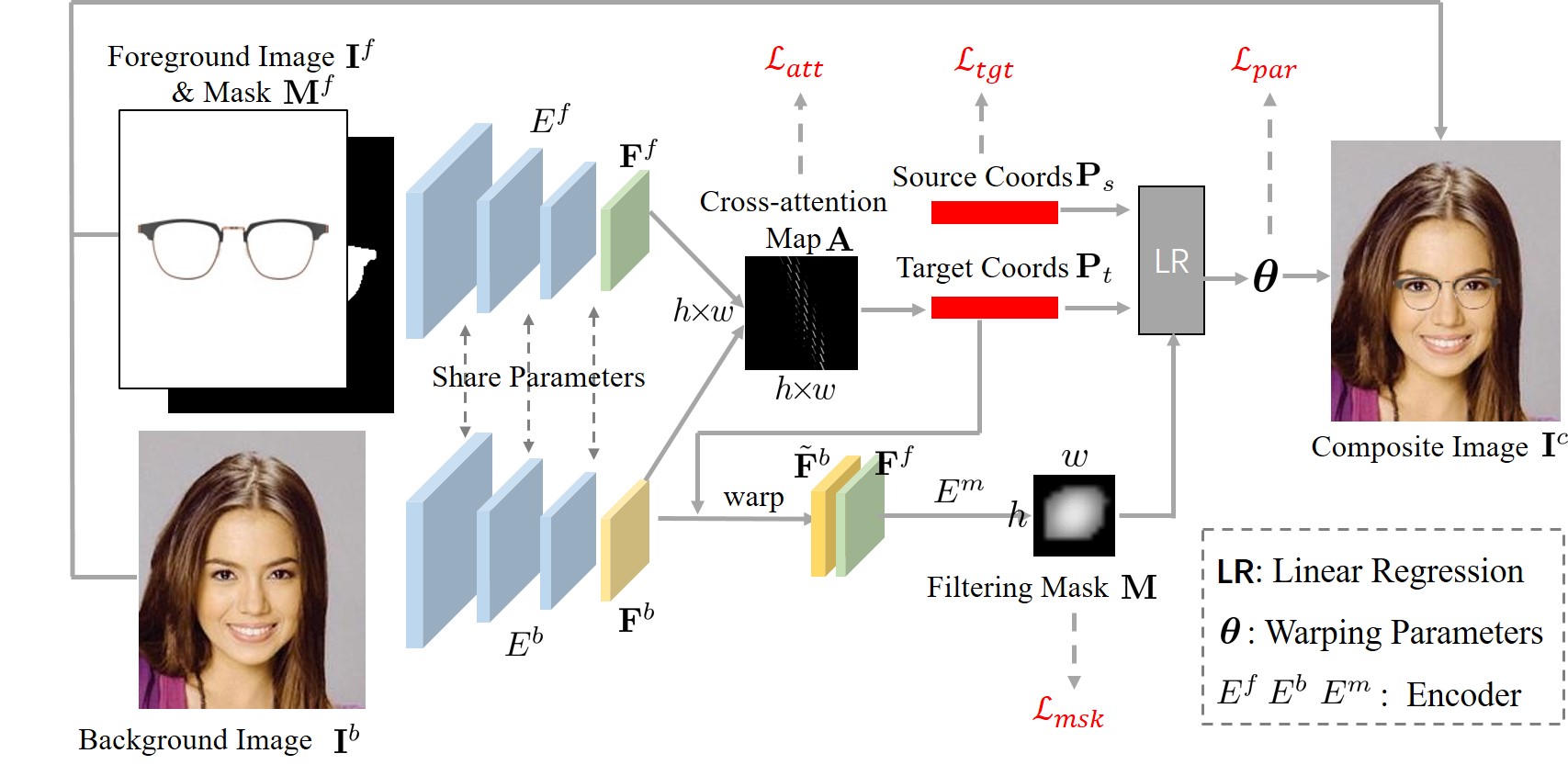}
    \caption{The illustration of our CorrelNet. Given a pair of foreground $\{\mathbf{I}^f, \mathbf{M}^f\}$ and background $\mathbf{I}^b$, we calculate their cross-attention map $\mathbf{A}$ to predict the target coordinates $\mathbf{P}_t$. Then, based on pairs of source coordinates $\mathbf{P}_s$ and target coordinates $\mathbf{P}_t$, we can solve the warping parameters $\bm{\theta}$ using linear regression. We also apply filtering mask $\mathbf{M}$ to eliminate noisy pairs of coordinates. Finally, we obtain the composite image $\mathbf{I}^c$ by combining the background and the warped foreground. }\label{fig:network}
\end{figure*}

\section{Dataset Construction} \label{sec:dataset}
Our STRAT dataset contains three subdatasets: STRAT-glasses, STRAT-hat, and STRAT-tie, which correspond to ``glasses try-on", ``hat try-on", and ``tie try-on" respectively. By taking STRAT-glasses subdataset as an example, we describe our way to collect the annotations. We first gather 30 glasses from Internet (\emph{resp.}, 3000 face images without glasses from CelebA dataset \cite{celeba}), which are divided into 20 training glasses (\emph{resp.}, 2000 training faces) and 10 test glasses (\emph{resp.}, 1000 test faces) without overlap. We randomly sample glasses from 20 training glasses to be composited with 2000 training faces to construct the training set. We construct the test set in the same way. Thus, the training (\emph{resp.}, test) set contains in total 2000 (\emph{resp.}, 1000) pairs of glasses and face images.  For each pair of glasses (foreground) and face image (background), as shown in Figure~\ref{fig:annotation}, we ask human annotators to paste the foreground on the background and adjust four vertices of foreground to make the composite image realistic. After adjustment, the coordinates of four vertices are recorded, based on which the ground-truth warping parameters can be calculated using linear regression (see Section \ref{sec:calculate_params}). 

For STRAT-hat, we also annotate 3000 pairs of hats and human faces, in which hats are collected from Internet and human face images without hats are collect from the same CelebA dataset \cite{celeba} as for STRAT-glasses. One critical issue is that some types of hats are not suitable for certain face poses, for example, a symmetrical baseball cap for leftward face pose. Such deformation cannot be modeled by simple perspective transformation. Thus, we only select compatible pairs for annotations. Another issue is that the hat may be out of the scope of face image. Our solution is boundary padding for the face image so that the entire hat can be included in the padded face image. For STRAT-tie, we consider two types of ties: bowtie and necktie. We collect 30 ties (15 neckties and 15 bowties) from Internet and 3000 portrait images without ties from Clothing Co-Parsing (CCP) dataset \cite{ccp} and Active Template Regression (ATR) dataset \cite{atr}. 

When annotating three subdatasets, we design some guiding rules (\emph{e.g.}, canonical location and scale of foreground,  correspondence between certain points in the foreground and background,  parallelism between certain lines in the foreground and background) to mitigate the uncertainty and make the annotation principles across different human annotators consistent. 
Some examples of foreground, background, and annotated composite image with composite foreground mask from three subdatasets are show in Figure \ref{fig:dataset}. 

\section{Problem Definition}
In the remainder of this paper, for ease of representation, we use a plain (\emph{resp.}, bold) letter $a$ (\emph{resp.}, $\mathbf{A}$) to denote a scalar (\emph{resp.}, vector or matrix). We use $\mathbf{A}^T$ to represent the transpose of matrix $\mathbf{A}$. Besides, we use $\mathbf{A}[i,:]$ (\emph{resp.}, $\mathbf{A}[:,j]$) to denote the $i$-th row (\emph{resp.}, $j$-th column) in  matrix $\mathbf{A}$, and 
$A[i,j]$ to denote the element in the $i$-row and $j$-th column in matrix $\mathbf{A}$. 

As illustrated in Figure~\ref{fig:task}, given a pair of foreground image $\mathbf{I}^{f}$ and background image $\mathbf{I}^{b}$, our goal is predicting the warping parameters $\bm{\theta}$ to perform geometric transformation for the foreground object. 
The foreground mask $\mathbf{M}^{f}$ indicates the foreground region in the foreground image $\mathbf{I}^{f}$.
Note that the foreground image and foreground mask would be warped synchronically. We use $\psi(\mathbf{I}^{f}, \bm{\theta})$ to denote the warped foreground by warping the foreground $\mathbf{I}^{f}$ with parameters $\bm{\theta}$. 
Then, the composite image $\mathbf{I}^{c}$ can be obtained by
\begin{align}\label{eqn:composite}
\mathbf{I}^c =\psi(\mathbf{I}^{f}, \bm{\theta}) \oplus \mathbf{I}^{b},
\end{align}
in which $\oplus$ indicates compositing the warped foreground and the background. The implementation details of $\psi(\cdot, \cdot)$ and $\oplus$ can be found in Supplementary.

\section{Our Method} \label{sec:method}
We employ two encoders $E^{f}$ and $E^{b}$ for foreground and background respectively. Specifically, the foreground encoder $E^f$ takes in the foreground image $\mathbf{I}^{f}$ and the foreground mask $\mathbf{M}^{f}$, and outputs the foreground feature map $\mathbf{F}^f\in\mathcal{R}^{h\times w \times c}$.
The background encoder $E^b$ takes in the background image  $\mathbf{I}^b$ and outputs the background feature map $\mathbf{F}^b \in \mathcal{R}^{h\times w \times c}$. The foreground and background share partial layers to reduce the number of parameters (see Figure \ref{fig:network}). With extracted foreground feature map and background feature map, we calculate their cross-attention map and predict the target locations for all foreground pixels, based on which the warping parameters $\bm{\theta}$ can be calculated. Moreover, we also design a filtering strategy to remove the noisy pairs of source and target locations, which helps achieve more accurate estimation of warping parameters. Next, we will elaborate on each step one by one. 

\subsection{Predicting Target Locations}
Based on the extracted foreground feature map $\mathbf{F}^f$ and background feature map $\mathbf{F}^b$, we calculate the their cross-attention map implying their  mutual correspondence, that is, where each foreground pixel should be moved to on the background. We project $\mathbf{F}^f$ and $\mathbf{F}^b$ into a common space using $\phi^f(\cdot)$ and $\phi^b(\cdot)$ respectively,  where $\phi^f(\cdot)$ and $\phi^b(\cdot)$  are $1 \times 1$ convolutional layer. For ease of calculation, we reshape $\phi^f(\mathbf{F}^f)\in \mathcal{R}^{h \times w \times c'}$ and  $\phi^b(\mathbf{F}^b)\in \mathcal{R}^{h \times w \times c'}$ to  $\bar{\phi}^f(\mathbf{F}^f)\in \mathcal{R}^{n \times c'}$ and  $\bar{\phi}^b(\mathbf{F}^b)\in \mathcal{R}^{n \times c'}$ respectively, in which $n=h\times w$ and $c'$ is the feature dimension after projection. Then, we calculate the cross-attention map as
\begin{eqnarray} \label{eqn:cross_attention}
\mathbf{A} = softmax(\bar{\phi}^f(\mathbf{F}^f)\bar{\phi}^b(\mathbf{F}^b)^T).
\end{eqnarray}

According to the ground-truth warping parameters, we can easily get the ground-truth target location of each foreground pixel.  The ground-truth binary cross-attention map for each foreground pixel can be acquired by setting the target location as $1$ and the other locations as $0$. 
For error tolerance and easy optimization, we apply Gaussian smooth with radius being 3 to the cross-attention map. By using $\mathbf{A}[i,:] \in \mathcal{R}^{n}$ (\emph{resp.}, $\hat{\mathbf{A}}[i,:]$ ) to denote the calculated (\emph{resp.}, ground-truth) cross-attention map for the $i$-th foreground pixel, the cross-attention map loss can be written as
\begin{eqnarray}\label{eqn:loss_att}
\mathcal{L}_{att} = \frac{1}{n} \sum^{n}_{i} \sum^{n}_{j} \alpha_{i,j} (A[i,j] - \hat{A}[i,j])^2, 
\end{eqnarray}
where $\alpha_{i,j}$ is a weighting factor introduced for addressing class imbalance. In our implementation, $\alpha_{i,j}$ is set as 1 for $A[i,j] < \hat{A}[i,j]$ and 0.25 for $A[i,j] \geq \hat{A}[i,j]$.

Intuitively, given $\mathbf{A}[i,:]$, we can obtain the target location of the $i$-th foreground pixel by selecting the location $j$ with the largest value $A[i,j]$ or averaging the locations with top largest values. However, in practice, we observe that it is difficult to obtain the accurate target location in this way. Thus, we adopt a learnable approach based on $\mathbf{A}[i,:]$ to predict more accurate target coordinate.
Formally, we assume that the coordinate of $i$-th foreground pixel is $\mathbf{p}^s_i=(x^s_i, y^s_i)$, which is deemed as the source location. The coordinate of its ground-truth target location on the background is $\hat{\mathbf{p}}^t_i=(\hat{x}^t_i, \hat{y}^t_i)$.  We apply a fully-connected layer to $\mathbf{A}[i,:]$ to predict the target coordinate $\mathbf{p}^t_i$, which is supervised by the ground-truth target coordinate:
\begin{eqnarray}\label{eqn:loss_tgt}
\mathcal{L}_{tgt} = \frac{1}{n} \sum^{n}_{i} \| \mathbf{p}^t_i - \hat{\mathbf{p}}^t_i \|_2^2. 
\end{eqnarray}

With predicted target locations $\{\mathbf{p}^t_i|_{i=1}^n\}$ for all foreground pixels, we can get $n$ pairs of source coordinates and target coordinates $\{(\mathbf{p}^s_i, \mathbf{p}^t_i)|_{i=1}^n\}$.

\subsection{Calculating Warping Parameters} \label{sec:calculate_params}
Based on $n$ pairs of source and target coordinates $\{(\mathbf{p}^s_i, \mathbf{p}^t_i)|_{i=1}^n\}$, we can have a close-form solution to the warping parameters $\bm{\theta} \in \mathcal{R}^{3 \times 3}$ using linear regression. Specifically, according to the definition of perspective transformation, the point-wise correspondence of $\mathbf{p}^s_i$ and $\mathbf{p}^t_i$ is projected by $\bm{\theta}$ using
\begin{equation} \label{eqn:perspective}
    \left[\begin{array}{c}
    x^{t}_i \\
    y^{t}_i \\
    1
    \end{array}\right] \sim\left[\begin{array}{llc}
    \theta_{11} & \theta_{12} & \theta_{13} \\
    \theta_{21} & \theta_{22} & \theta_{23} \\
    \theta_{31} & \theta_{32} & \theta_{33}
    \end{array}\right]\left[\begin{array}{c}
    x^{s}_i \\
    y^{s}_i \\
    1
\end{array}\right] \Leftrightarrow \mathbf{P}^{T}_{t} \sim \bm{\theta} \mathbf{P}^{T}_{s}, 
\end{equation}
where $\mathbf{P}_{s}, \mathbf{P}_{t} \in \mathcal{R}^{n \times 3}$ denote the source and target coordinate matrices using homogeneous coordinates respectively.  
Given that the number of paired points far exceeds the number of unknown parameters, we solve for the warping parameters $\bm{\theta}$ by minimizing the following objective and obtain the close-form solution to $\bm{\theta}$:
\begin{equation} \label{eqn:param_solve}
    \min _{\bm{\theta}} \|\mathbf{P}^T_t -\bm{\theta} \mathbf{P}^T_s\|^{2}_{2} \Rightarrow 
    \bm{\theta}= \mathbf{P}_t^T \mathbf{P}_s \left(\mathbf{P}_s^T \mathbf{P}_s\right)^{-1}.
\end{equation}

Similarly, given four pairs of source coordinates and ground-truth target coordinates as shown in Figure \ref{fig:annotation}, we can obtain the ground-truth warping parameters $\hat{\bm{\theta}}$ according to Eqn. (\ref{eqn:param_solve}).
During training phase, the predicted warping parameters $\bm{\theta}$ are supervised with ground-truth warping parameters $\hat{\bm{\theta}}$ by a smooth L1 loss \cite{ren2015faster} considering its robustness to outliers:
\begin{equation} \label{eqn:param_loss}
    \mathcal{L}_{par} = \frac{1}{9} \sum^{3}_{i} \sum^{3}_{j} \mathcal{L}_{s1}(\theta_{i,j} - \hat{\theta}_{i,j}),
\end{equation}
where $\mathcal{L}_{s1}(\cdot)$ represents the smooth L1 loss
\begin{equation}
    \mathcal{L}_{s1}(x)=\left\{\begin{array}{cc}
                0.5 x^{2}, & \mathrm{if}\ x<1, \\
                |x|-0.5, &   \mathrm{otherwise}.
                \end{array}\right.
\end{equation}
With the predicted warping parameters $\bm{\theta}$, we can warp the foreground and generate the composite image using Eqn. (\ref{eqn:composite}). 

\subsection{Filtering Strategy} \label{sec:filtering_strategy}
One remaining problem is that if we solve the linear regression problem based on all $n$ pairs, some noisy and unreliable pairs may mislead the solution, leading to inaccurate warping parameters. To suppress the negative impact of noisy pairs, we attempt to predict a filtering mask to discard noisy pairs. 

We first align the background feature map $\mathbf{F}^b$ with foreground feature map $\mathbf{F}^f$. Specifically, according to the coordinate pairs $\{(\mathbf{p}^s_i, \mathbf{p}^t_i)|_{i=1}^n\}$, we can warp the background feature map as $\tilde{\mathbf{F}}^b$ with the $i$-th element $\tilde{\mathbf{F}}^b[x^s_i, y^s_i]$  being $\mathbf{F}^b[x^t_i, y^t_i]$.
Then, we concatenate foreground feature map $\mathbf{F}^f$ and aligned background feature map $\tilde{\mathbf{F}}^b$, which passes through the filtering mask encoder $E^{m}$ to predict the filtering mask $\mathbf{M} \in \mathcal{R}^{h \times w}$ with entry values in $[0,1]$. The intuition is that by comparing the aligned pixel-wise features, the model is expected to identify those misaligned pixels corresponding to noisy pairs. 
We expect the filtering mask $\mathbf{M}$ to be negatively correlated with the target prediction error $\|\mathbf{p}^t_i-\hat{\mathbf{p}}^t_i\|_1$. In other words, the larger the target prediction error, the smaller the filter mask value. We use $\mathbf{E}\in \mathcal{R}^{h \times w}$ to denote the target prediction error map. 
Then, we flatten $\mathbf{M}$ (\emph{resp.}, $\mathbf{E}$) to $\bar{\mathbf{M}}\in \mathcal{R}^{n}$ (\emph{resp.}, $\bar{\mathbf{E}}\in \mathcal{R}^{n}$) and use $\bar{M}_{i}$ (\emph{resp.}, $\bar{E}_{i}$) to denote the $i$-th entry in $\bar{\mathbf{M}}$ (\emph{resp.}, $\bar{\mathbf{E}}$), in which $\bar{E}_i=\|\mathbf{p}^t_i-\hat{\mathbf{p}}^t_i\|_1$.
The correlation between  $\bar{\mathbf{M}}$ and $\bar{\mathbf{E}}$ is calculated as follows,
\begin{eqnarray}\label{eqn:loss_filtering}
\mathcal{L}_{msk} =\frac{\sum_{i} (\bar{M}_{i} - \bar{m})(\bar{E}_{i} - \bar{e}) }{\| \sum_{i} (\bar{M}_{i} - \bar{m})^2 \sum_{i} (\bar{E}_{i} - \bar{e})^2 \|^{\frac{1}{2}}},
\end{eqnarray}
where $\bar{m}$ and $\bar{e}$ are the average values of $\bar{\mathbf{M}}$ and $\bar{\mathbf{E}}$, respectively. By minimizing the correlation in Eqn. (\ref{eqn:loss_filtering}), we expect the filtering mask to be negatively correlated with the target prediction error. 

On the premise of the filtering mask $\mathbf{M}$, we keep the pairs with values above the average value $\bar{m}$ and calculate the warping parameters according to Eqn. (\ref{eqn:param_solve}). Finally, we warp the foreground and composite it with the background using Eqn. (\ref{eqn:composite}), yielding the composite image $\mathbf{I}^c$. 
The overall optimization function can be written as
\begin{equation}
\label{eqn:total_loss}
    \mathcal{L} = \mathcal{L}_{tgt} + \lambda_{att} \mathcal{L}_{att} + \lambda_{par} \mathcal{L}_{par} + 
    \lambda_{msk} \mathcal{L}_{msk},
\end{equation}
where $\lambda_{att}$, $\lambda_{par}$, and $\lambda_{msk}$ are trade-off parameters.

\section{Experiments}

\begin{table*}
\setlength{\tabcolsep}{1.mm}
\begin{tabular}{l|cccc|cccc|cccc}
\hline
\multirow{2}*{Method} & \multicolumn{4}{c|}{STRAT-glasses} &\multicolumn{4}{c|}{STRAT-hat} &\multicolumn{4}{c}{STRAT-tie}\\
& LSSIM↑   & IoU↑  & Disp↓   &User↑ & LSSIM↑  & IoU↑  & Disp↓ &User↑ & LSSIM↑  & IoU↑  & Disp↓ &User↑\\ 
                         \hline \hline
ST-GAN  \cite{stgan} & 0.5655  & 0.5932 & 0.0240  & 7.1\% & 0.4362 & 0.6859 & 0.0316  & 9.0\% & 0.2780 & 0.1126 & 0.0440  & 8.4\% \\
ST-GAN (+s) \cite{stgan} & 0.6061 & 0.6579 & 0.0198  & 16.8\% & 0.5164 	& 0.7455  & 0.0235   & 18.2\% & 0.2517 & 0.1211 & 0.0393  & 11.3\% \\
CompGAN \cite{Compositional-GAN} & 0.5362 & 0.5593 & 0.0279  & 5.2\% & 0.4311 & 0.6675 & 0.0343  & 7.4\% & 0.2768 	& 0.0918 & 0.0506  & 5.5\% \\
CompGAN (+s) \cite{Compositional-GAN} & 0.5807 	& 0.6353 & 0.0216  & 13.8\% & 0.5047  & 0.7303  & 0.0246  & 16.5\% & 0.2452 	& 0.1064 & 0.0425   & 6.7\% \\
RegGAN \cite{RegGAN2019} & 0.5356 & 0.5069 & 0.0299 & 3.3\% & 0.4028 & 0.6147 & 0.0371  & 4.5\% & 0.2469 	& 0.0792 	& 0.0603  & 2.8\%\\
SF-GAN \cite{SF-GAN} & 0.5406 & 0.5472 & 0.0267  & 4.9\% & 0.4140 & 0.6521  & 0.0365  & 5.7\% & 0.2575 	& 0.0885 & 0.0544  & 4.4\%\\
AGCP \cite{Li2021ImageSV} & 0.5240 	& 0.4750 & 0.0347  & 1.4\% & 0.3954  & 0.5981 & 0.0386  & 2.7\% & 0.2348 & 0.0641 	& 0.0640  & 0.7\% \\
Ours & \textbf{0.6886} & \textbf{0.7573} & \textbf{0.0145} & \textbf{47.5\%} 
     & \textbf{0.5470} & \textbf{0.7873} & \textbf{0.0184} & \textbf{36.0\%}
     & \textbf{0.2883} & \textbf{0.3948} & \textbf{0.0131} & \textbf{60.2\%} \\ \hline
\end{tabular}
\caption{Results of quantitative comparison on our STRAT dataset. ``User'' means the frequency that each method is chosen as the best method in user study.}
\label{tab:baseline}
\end{table*}

\subsection{Dataset and Evaluation Metrics}
As described in Section \ref{sec:dataset}, our STRAT dataset consists of three subdatasets. In each subdataset, the training set has 2000 pairs of foregrounds and backgrounds, while the test set has 1000 pairs of foregrounds and backgrounds. The foreground/background images in the training set and test set have no overlap. 

Since this is the first dataset with ground-truth annotations for this task, we deliberately design three evaluation metrics to measure the discrepancy between generated composite images and ground-truth composite images. 
\begin{itemize}
    \item LSSIM: The structural similarity index (SSIM) \cite{Wang2004ImageQA} is a commonly used metric to measure the difference between two images. Considering that the foreground may only occupy a small area in the whole image, we calculate SSIM in a local region named Local SSIM (LSSIM), which focuses on the foreground region. Specifically, we crop the bounding box for the foreground in the ground-truth composite image and calculate SSIM within this bounding box.   
    \item Disp: Inspired by previous works \cite{Le2020Homography,Chang2017CLKN,DeTone2016DeepIH} related to spatial transformation, we calculate the vertex displacement error (Disp) between predicted four vertices and ground-truth four vertices, which measures their L1 distance normalized by the image size and then takes average over the four vertices. As shown in Figure \ref{fig:annotation}, A', B', C', and D' are ground-truth four vertices. 
    \item IoU: Inspired by segmentation evaluation, we adopt intersection over union (IoU) as another metric. Precisely, we calcualte IoU between the predicted composite foreground mask and the ground-truth composite foreground mask. 
\end{itemize}

Considering that the above quantitative metrics sometimes may not conform with human perception, we also include user study for more comprehensive comparison. Specifically, we first collect the composite images generated by different methods for the whole test set, and then ask 20 human raters to select the best composite image for each testing pair. Next, we count the frequency that each method is chosen as the best method and present the results in Table \ref{tab:baseline}.

\subsection{Implementation Details}
We use PyTorch \cite{paszke2019pytorch} to implement our model, which is distributed on NVIDIA RTX 3090 GPU. When training and testing on START-glasses and STRAT-hat subdatasets, all background images and foreground images/masks are resized to $224 \times 224$. For STRAT-tie subdataset, considering the small foreground area on composite image, the input image and mask are resized to $448 \times 448$ for higher resolution.
We use the first three stages of the ResNet-50 \cite{DBLP:conf/cvpr/HeZRS16} pretrained on ImageNet \cite{deng2009imagenet} as the backbone for both background and foreground encoders $\{E^b, E^f\}$. To save model parameters, two encoders share all layers except the first convolutional layer, due to different input channel dimensions. When the input size is $224 \times 224$, the foreground and background feature maps have the same shape of $14 \times 14 \times 1024$, \emph{i.e.}, $h=w=14, c=1024$. The feature dimension after projection is $c'=512$.  
We implement the filtering mask encoder $E^{m}$ by employing two convolution+relu operations, followed by a fully-connected layer and a $sigmoid$ function.
We use adam optimizer with the learning rate initialized as 0.0002 and $\beta$ set to (0.9, 0.999). The batch size is 1 and our model is trained for 20 epochs.
We set $\lambda_{att}=1$, $\lambda_{par}=0.01$, and $\lambda_{msk}=0.1$ via cross-validation. The results using different hyper parameters and backbones are left to Supplementary.

\subsection{Comparison with Baselines}
We compare with previous methods \cite{stgan,RegGAN2019}, which focus on spatial transformation for image composition. We also compare with \cite{Compositional-GAN,SF-GAN,Li2021ImageSV} which deal with both appearance and geometry inconsistency between foreground and background. In other words, they not only learn the warping parameters for the foreground object, but also adjust the appearance of foreground object. For these methods \cite{Compositional-GAN,SF-GAN,Li2021ImageSV}, we remove the modules related to appearance adjustment and only keep the modules related to spatial transformation. 

\begin{figure}[t]
    \centering
    \includegraphics[width=0.9\linewidth]{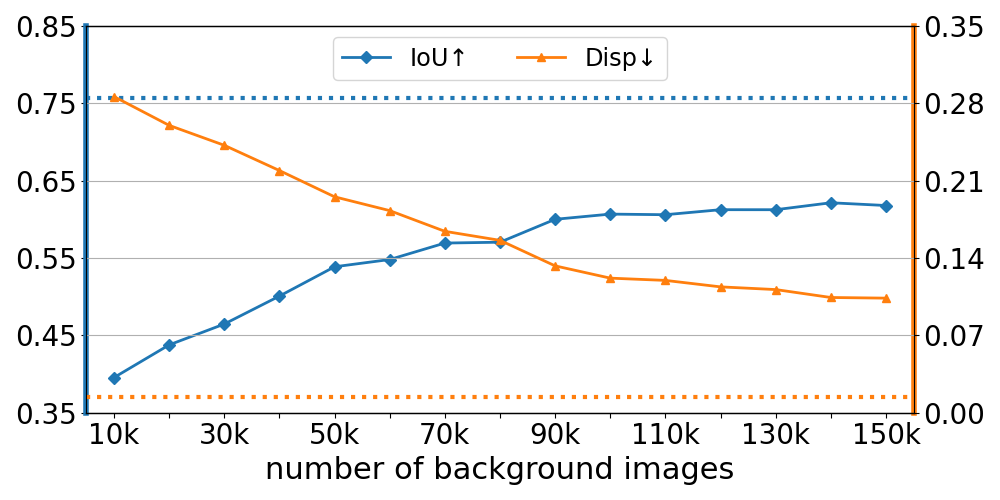}
    \caption{The results of ST-GAN \cite{stgan} training with different numbers of background images following the setting in \cite{stgan}. The dashed horizontal lines indicate the results of our model.}
    \label{fig:different_bg}
\end{figure}

For fair comparison, we adopt the ResNet-50 \cite{DBLP:conf/cvpr/HeZRS16} pretrained on ImageNet \cite{deng2009imagenet} as the backbone for baseline methods. We also use the same training set for all baseline methods. In particular, we use training pairs of foregrounds and backgrounds as input, while using annotated composite images as positive images.
It is worth mentioning that although the above methods can utilize annotated composite images (\emph{i.e.}, using these positive samples when updating the discriminator), they do not directly use the ground-truth annotations and thus cannot release the full potential of annotations. Therefore, we select two competitive baselines (\emph{i.e.}, ST-GAN \cite{stgan} and CompGAN \cite{Compositional-GAN}) and adapt them to supervised version. Specifically, apart from using positive samples to update the discriminator, we also force their predicted composite images to be close to the ground-truth composite images using mean square error (MSE) loss. We refer to the modified version as ``ST-GAN (+s)'' and ``CompGAN (+s)'', which means adding direct supervision. 

The results of baselines and our CorrelNet are summarized in Table \ref{tab:baseline}. It can be seen that among the original baselines, ST-GAN \cite{stgan} is the most competitive baseline, probably because ST-GAN employs multiple generators to iteratively update the warping parameters. 
By comparing the original version ST-GAN (\emph{resp.}, CompGAN) and modified version ST-GAN (+s) (\emph{resp.}, CompGAN (+s)), we can see that the modified version generally achieves better performance in terms of IoU and Disp, which demonstrates the effectiveness of directly using annotations. Nevertheless, there is still a clear performance gap between ST-GAN (+s) and our model, which implies that adding supervision to the composite image is insufficient.     
Among three subdatasets, we find that the performances on STRAT-tie subdataset are worse than the other two subdatasets. One possible explanation is that the bowties on STRAT-tie subdataset are relatively small and the neckties generally have an extreme aspect ratio, both of which make it challenging to warp them properly to background. Despite these challenges, our CorrelNet beats all the baseline methods by a large margin for all evaluation metrics on all three subdatasets. 

In the original setting in previous works \cite{stgan,Compositional-GAN,SF-GAN}, they only utilize unannotated data. For example, ST-GAN \cite{stgan} uses pairs of glasses (foreground) and face image without glasses from CelebA dataset \cite{celeba} (background) as input, and adopts the face images with glasses from CelebA dataset as positive samples to update the discriminator. In this way, they can use all 152249  face images without glasses from CelebA dataset as backgrounds. Here, we following the setting in \cite{stgan} and use different numbers ([10k, 20k, ..., 150k]) of background images. We plot the results using different numbers of background images in Figure \ref{fig:different_bg}, from which we observe that the performance first increases and then begins to converge when the number of background images exceeds 90k. However, the optimal performance using 150k background images is still worse than our method, and even worse than ST-GAN in Table \ref{tab:baseline} in terms of Disp. We conjecture that for ST-GAN in Table \ref{tab:baseline}, the input foreground-background pairs are coupled with the positive samples, that is, the positive composite images are created based on input foregrounds/backgrounds, which can provide more effective supervision information. 

\subsection{Ablation Studies} \label{sec:ablation_study}
\begin{table}
\centering
\setlength{\tabcolsep}{1.2mm}
\begin{tabular}{l|cccc|ccc}
\hline
& $\mathcal{L}_{par}$  & $\mathbf{p}^t+\mathcal{L}_{tgt}$ & $\mathbf{A}$ & $\mathbf{M}$ & LSSIM↑   & IoU↑  & Disp↓ \\
\hline \hline
1 & +  &    &   &    & 0.5793          & 0.0031          & 0.2029          \\
2 &    & +  &   &    & 0.5564          & 0.6471          & 0.0192          \\
3 & +  & +  &   &    & 0.5905          & 0.6865          & 0.0192          \\
4 &    & +  & + &    & 0.6431	       & 0.7161	         & 0.0177          \\
5 & +  & +  & + &    & 0.6654          & 0.7234          & 0.0161          \\
6 & +  &max & + &    & 0.4909	       & 0.5032	         & 0.0238          \\
7 & +  &avg & + &    & 0.4994	       & 0.5331	         & 0.0229          \\
8 & +  & +  & + & fm & 0.5696	       & 0.4771	         & 0.0276          \\
9 & +  & +  & + & ff & 0.6682	       & 0.7355	         & 0.0153          \\
10 & + & +  & + & +  & \textbf{0.6886} & \textbf{0.7573} & \textbf{0.0145} \\
\hline
\end{tabular}
\caption{Ablation studies of loss terms and alternative network designs of our CorrelNet.
``max'' and ``avg'' are two ways to obtain the target locations based on cross-attention map. ``fm'' and ``ff'' are short for ``foreground mask" and ``foreground feature" respectively, which represent two ways to predict the filtering mask. The detailed explanations can be found in Section~\ref{sec:ablation_study}.
}
\label{tab:ablation_study}
\end{table}
To investigate the effectiveness of each component in our CorrelNet, we conduct ablation studies in Table \ref{tab:ablation_study}. First, we directly concatenate foreground feature map $\mathbf{F}^f$ and background feature map $\mathbf{F}^b$ to predict the warping parameters without calculating cross-attention map, during which only the warping parameters loss $\mathcal{L}_{par}$ is used. As reported in row 1, the performance is very poor, which implies that directly adding supervision to the warping parameters is ineffective. Then, we use the concatenation of feature maps to predict target coordinates $\mathbf{p}^t$ by using $\mathcal{L}_{tgt}$. The results are reported in row 2, which are far better than row 1 in terms of IoU and Disp. This demonstrates that predicting coordinates is much more effective than predicting warping parameters. Based on row 2, we further add $\mathcal{L}_{par}$ and achieve better results in row 3, which shows that $\mathcal{L}_{par}$  is complementary with $\mathcal{L}_{tgt}$ and can help improve the performance.  

Next, we calculate cross-attention map as in our method and adopt the losses $\{\mathcal{L}_{tgt}, \mathcal{L}_{att}\}$, which is reported in row 4. We find that introducing cross-attention map can bring notable performance gain. Based on row 4, we further add $\mathcal{L}_{par}$ and the achieved results in row 5 again verify the effectiveness of $\mathcal{L}_{par}$.  Then, we explore several variants of target coordinate prediction. 
First, we try acquiring the target coordinate based on cross-attention map $\mathbf{A}$ without learning process. One simple approach is taking the location with the largest value in $\mathbf{A}[i,:]$, \emph{i.e.}, $j=\arg\max_j A[i,j]$, as the target location for the $i$-th foreground pixel. The obtained results in row 6 become worse. We also calculate the weighted average coordinate, \emph{i.e.}, $j=\frac{\sum A[i,j]*j}{\sum A[i,j]}$ (row 7). The results in row 7 are slightly better than row 6 but still quite disappointing. The results in row 6 and 7 demonstrate that although the cross-attention map can roughly capture the correspondence, it is still unreliable to directly acquire the target coordinates without learning process. 

Based on row 5, we employ filtering mask to remove noisy pairs and arrive at our full-fledged method in row 10. The advantage of row 10 over row 5 proves the effectiveness of filtering out noisy pairs. Then, we explore two variants of filtering mask. Firstly, we directly use the foreground mask as filtering mask without learning process (row 8), from which we observe significant performance drop. Secondly, we only use the foreground feature map $\mathbf{F}^f$ to predict the filtering mask (row 9). The results in row 9 are worse than our full method in row 10, indicating the necessity of using both foreground feature map and aligned background feature map to predict the filtering mask. 
Apart from quantitative comparison, we also show some visualization examples for ablation study in Supplementary.  

\subsection{Qualitative Analyses}

\subsubsection{Cross-attention Map}
\begin{figure}[t]
    \centering
    \includegraphics[width=0.9\linewidth]{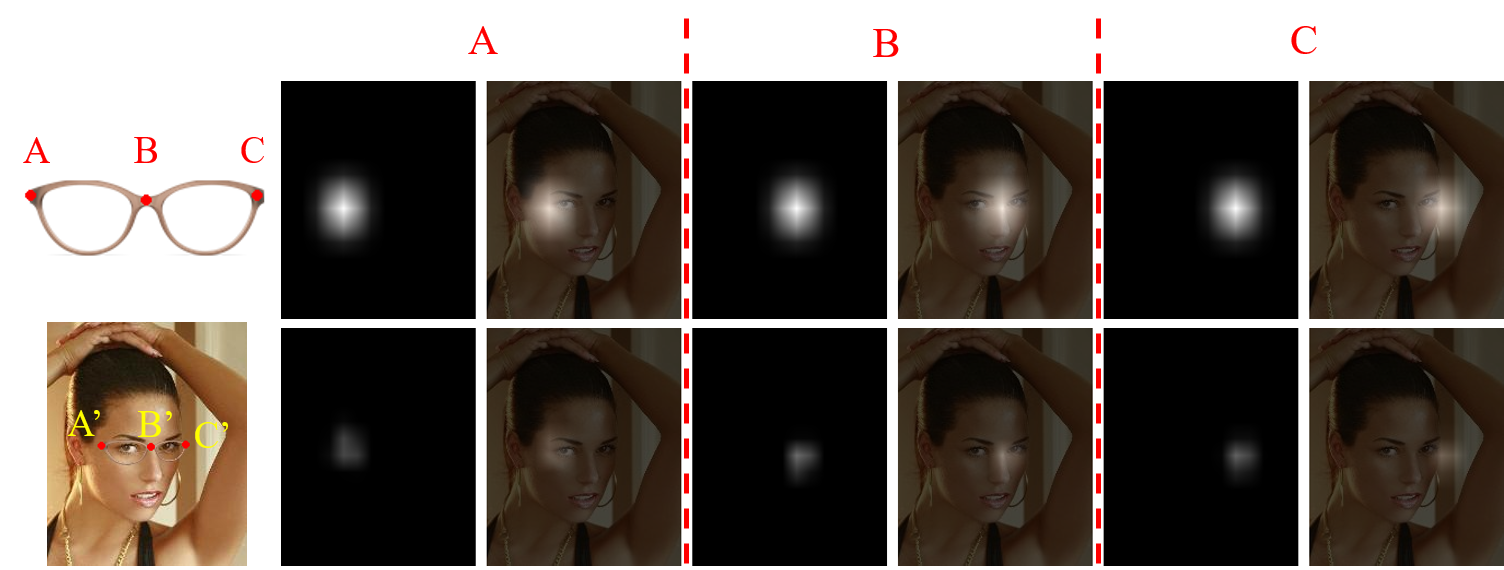}
    \caption{Visualization of the predicted (\emph{resp.}, ground-truth) cross-attention map in the bottom (\emph{resp.}, top) row for three key points (A, B, C). We also superimpose attention map on the background for better visualization.}\label{fig:cross_attention}
\end{figure}

We visualize the cross-attention map $\mathbf{A}\in\mathcal{R}^{n\times n}$ with $n=w\times h$. For ease of visualization, we select some key points $i$ from the foreground image and reshape the corresponding row in the cross-attention map $\mathbf{A}[i,:]\in \mathcal{R}^{n}$ to the size $w\times h$. As shown in Figure \ref{fig:cross_attention}, we mark three keypoints (A, B, C) on the foreground image and show their corresponding rows in the cross-attention map, from which we can see that the bright region in the cross-attention map can roughly match the expected target location on the background (\emph{e.g.}, the center of glasses matches the nose on the human face) and the predicted cross-attention map is close to the ground-truth cross-attention map. 

\subsubsection{Filtering Mask}
\begin{figure}[t]
    \centering
    \includegraphics[width=0.9\linewidth]{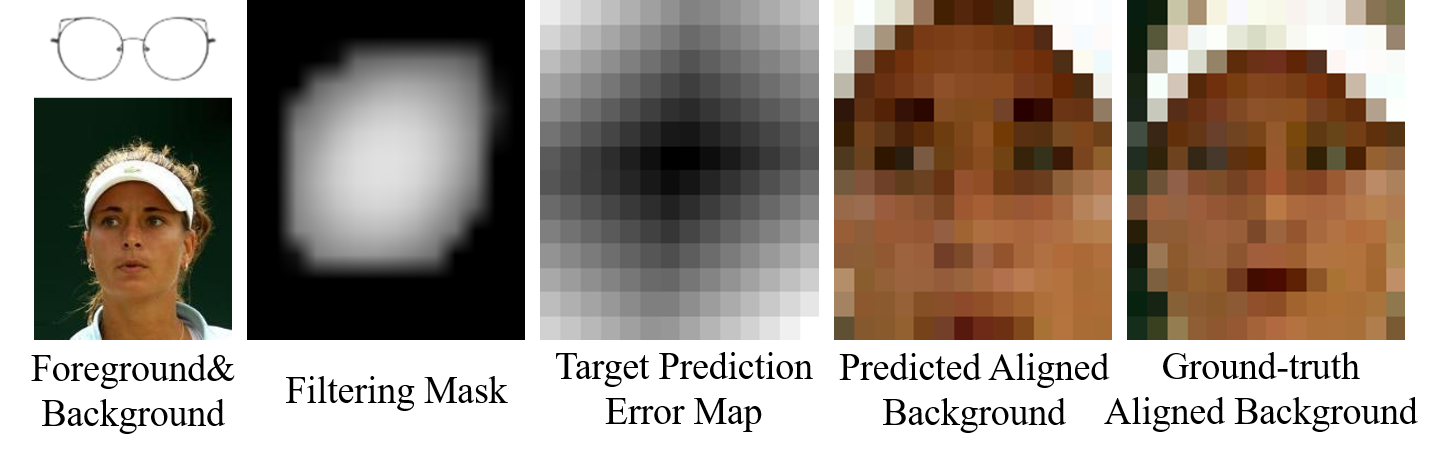}
    \caption{Visualization of predicted filtering mask, target prediction error map, the background images aligned by the predicted/ground-truth target coordinates.
    }\label{fig:filtering_mask}
\end{figure}

We visualize the predicted filtering mask $\mathbf{M}$ in Figure \ref{fig:filtering_mask}. Recall that the filtering mask $\mathbf{M}$ is expected to be negatively correlated with the target prediction error map $\mathbf{E}$, aiming to filter out noisy pairs of source and target coordinates (see Section~\ref{sec:filtering_strategy}). So we also show the target prediction error map $\mathbf{E}$ obtained by normalizing the L1 distance between the predicted and ground-truth target coordinates over all locations.
As demonstrated in Figure \ref{fig:filtering_mask}, it is obvious that $\mathbf{M}$ is negatively correlated with $\mathbf{E}$. Since larger target prediction error means larger misalignment, for intuitive illustration, we also show the aligned background image $\tilde{\mathbf{I}}^b$ by assigning the pixel values at target coordinates to the source coordinates, in a similar way to $\tilde{\mathbf{F}}^b$ (see Section~\ref{sec:filtering_strategy}). By comparing the predicted and ground-truth aligned background image, we can identify the regions with more notable misalignment, which match the dark regions in the filtering mask. 
This demonstrates that our predicted filtering mask can successfully remove the noisy pairs for more accurate estimation of warping parameters. 

\subsubsection{Composite Results}
\begin{figure}[t]
    \centering
    \includegraphics[width=1\linewidth]{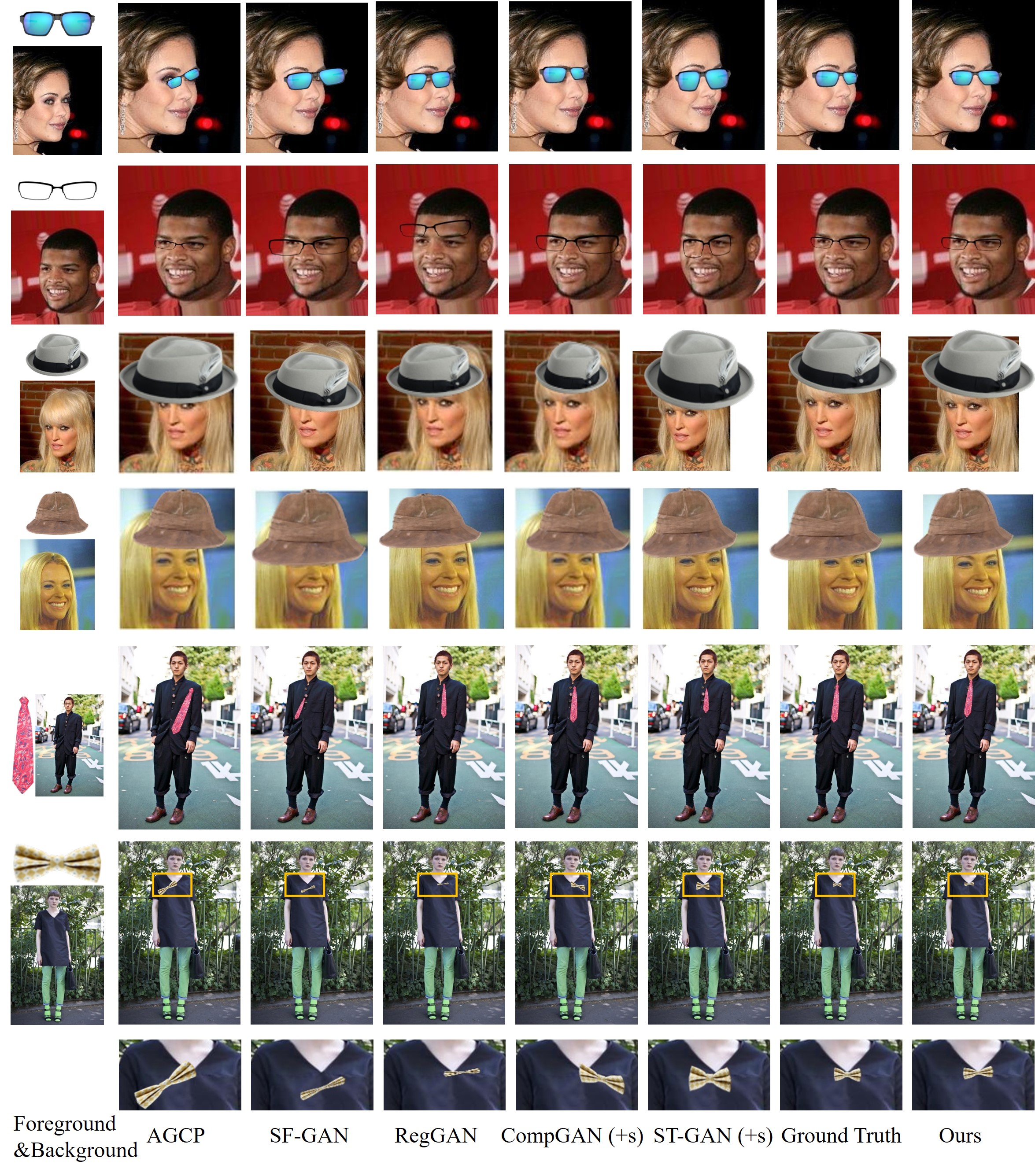}
    \caption{Qualitative comparison on our STRAT dataset. In the last row, we zoom in for clearer observation. 
    }\label{fig:qualitative_comparison}
\end{figure}
We show the resultant composite images of different methods on three subdatasets in Figure \ref{fig:qualitative_comparison}. Since the foreground bowties in STRAT-tie are very small, we zoom in the yellow bounding box for better observation. The results show that the baseline methods may warp foreground objects to unreasonable positions or shapes, resulting in implausible composite images.
In contrast, our CorrelNet can produce more realistic and plausible composite images, which are closer to ground-truth composite images. More visualization examples of cross-attention map, filtering mask, qualitative results, and the discussion on the limitation of our method are present in Supplementary.

\section{Conclusion}
In this work, we have contributed a STRAT dataset composed of three subdatasets: STRAT-glasses, STRAT-hat, STRAT-tie. We have also proposed a novel CorrelNet which predicts the target coordinate of foreground pixels based on the cross-attention map between foreground and background. This novel solution performs much better than directly predicting the warping parameters as in previous methods, which provides a new perspective for this task. 
We hope that our contributed dataset and proposed solution can shed light on the future research of spatial transformation for image composition.

{\small
\bibliographystyle{ieee_fullname}
\bibliography{egbib}
}

\end{document}


\title{Supplementary for Spatial Transformation for Image Composition via Correspondence Learning}

\author{$\textnormal{Bo Zhang}^{*}$,
$\textnormal{Yue Liu}^{*}$,
$\textnormal{Kaixin Lu}^{\dag}$,
$\textnormal{Li Niu}^{*}$, 
$\textnormal{Liqing Zhang}^{*}$\\
$^*$ Shanghai Jiao Tong University\,\,
$\dag$ Shanghai University \\
}
\maketitle

In this document, we provide additional materials to supplement our main text. In Section~\ref{sec:composition_details}, we provide more implementation details on image composition operation. Then, we study the impact of backbone network in Section~\ref{sec:backbone} and the effect of three hyper-parameters adopted in our loss function in Section~\ref{sec:hyper_param}. In Section~\ref{sec:abltation_study}, we show some example images produced by the ablated versions of our CorrelNet on our STRAT dataset. In Section~\ref{sec:qualitative_results}, we show more qualitative results on STRAT dataset, including the cross-attention map and filtering mask generated by our method, and composite results produced by different methods. In Section~\ref{sec:limitation}, we show some failure cases generated by our method and discuss the limitation of our method. 

\section{Image Composition Details} 
\label{sec:composition_details}
The goal of image composition is warping the foreground and then combining with the background to produce a composite image.
In Section 4 of the main text, we briefly describe the image composition operation  omitting some trivial details. Here, we will present a more detailed implementation.

Formally, given a pair of RGB background image $\mathbf{I}^{b} \in \mathcal{R}^{h \times w \times 3}$ and foreground image $\mathbf{I}^{f} \in \mathcal{R}^{h \times w \times 3}$ with mask $\mathbf{M}^{f} \in \mathcal{R}^{h \times w}$ that indicates the foreground region with entry values in $[0,1]$, we first use ground-truth or predicted warping parameters $\bm{\theta} \in \mathcal{R}^{3 \times 3}$ to perform perspective transformation on the foreground image $\mathbf{I}^{f}$ and foreground mask $\mathbf{M}^{f}$ synchronically, respectively yielding the warped foreground image $\tilde{\mathbf{I}}^{f}$ and warped foreground mask $\tilde{\mathbf{M}}^{f}$. Specifically, we assume that there are a total of  $n=h \times w$ pixels in the foreground image. The coordinate of $i$-th pixel on the foreground is denoted by $\mathbf{p}^s_i=(x^s_i, y^s_i)$, which is deemed as the source coordinate, and its coordinate on the warped foreground is $\mathbf{p}^t_i=(x^t_i, y^t_i)$, which is regarded as the target coordinate. The point-wise correspondence between $\mathbf{p}^s_i$ and $\mathbf{p}^t_i$ is realized by the projection matrix $\bm{\theta}$ using Eqn. (5) of the main text. For ease of implementation, we derive the source coordinate $\mathbf{p}^s_i$ for each target coordinate $\mathbf{p}^t_i$ of the regular grid in the warped foreground image by
\begin{equation} \label{eqn:perspective}
    \left[\begin{array}{llc}
    \theta_{11} & \theta_{12} & \theta_{13} \\
    \theta_{21} & \theta_{22} & \theta_{23} \\
    \theta_{31} & \theta_{32} & \theta_{33}
    \end{array}\right]^{-1}\left[\begin{array}{c}
    x^{t}_i \\
    y^{t}_i \\
    1
    \end{array}\right] \sim \left[\begin{array}{c}
    x^{s}_i \\
    y^{s}_i \\
    1
\end{array}\right],
\end{equation}
in which $x^t_i$ and $y^t_i$ are integers, while $x^s_i$ and $y^s_i$ are typically decimals.

\begin{figure}[t]
    \centering
    \includegraphics[width=1\linewidth]{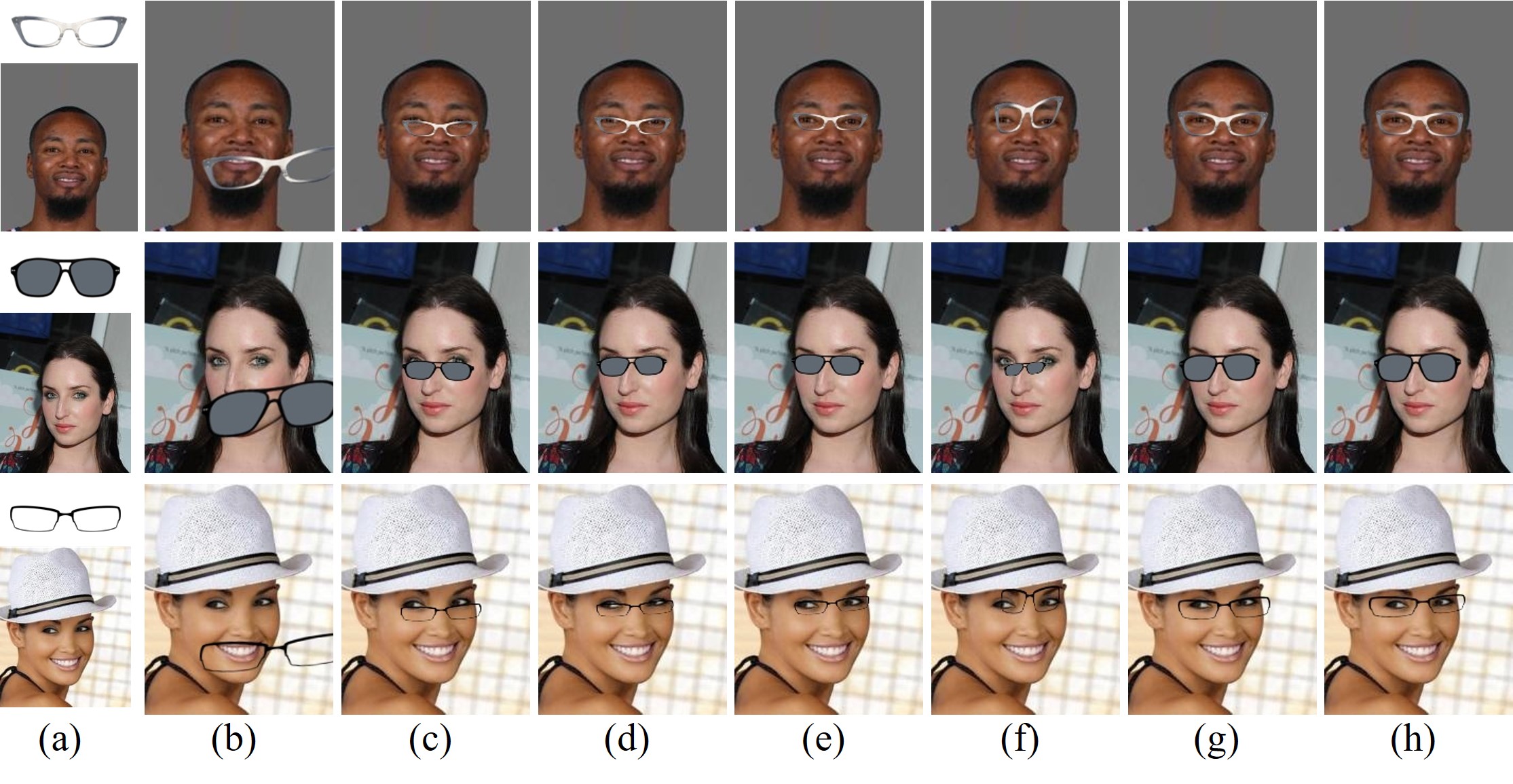}
    \caption{Visualization results of ablated versions of our CorrelNet on STRAT dataset. From left to right are foreground and background images (a), composite image produced by the ablated methods in row 1 (b), row 2 (c), row 3 (d), row 5 (e), row 8 (f) of Table 2 in the main text, our full-fledged method (g), ground-truth composite image (h). 
    }\label{fig:ablation_study}
\end{figure}

After calculating the coordinate pairs $\{(\mathbf{p}^s_i, \mathbf{p}^t_i)|_{i=1}^n\}$, we can warp the foreground image $\mathbf{I}^f$ as $\tilde{\mathbf{I}}^{f}$ with the $i$-th element $\tilde{\mathbf{I}}^f[x^t_i, y^t_i]$  being $\mathbf{I}^{f}[x^s_i, y^s_i]$, in which $\mathbf{I}^f[x^s_i, y^s_i]$ can be generated by bilinear interpolation. 
The foreground mask $\mathbf{M}^{f}$ is warped as $\tilde{\mathbf{M}}^{f}$ with the $i$-th element $\tilde{\mathbf{M}}^f[x^t_i, y^t_i]$ being $\mathbf{M}^{f}[x^s_i,y^s_i]$, in a similar way to $\tilde{\mathbf{I}}^{f}$.
For the source coordinate $(x^s_i, y^s_i)$ beyond the boundaries of foreground image $\mathbf{I}^{f}$, we fill the pixel values at the corresponding target coordinate $(x^t_i, y^t_i)$ with zeros, \emph{i.e.}, $\tilde{\mathbf{I}}^f[x^t_i, y^t_i]=\mathbf{0}$, $\tilde{\mathbf{M}}^f[x^t_i, y^t_i]=0$.

With background image $\mathbf{I}^b$, warped foreground image $\tilde{\mathbf{I}}^f$, and warped foreground mask $\tilde{\mathbf{M}}^f$ with entry values in [0,1], the composite image $\mathbf{I}^{c}$ can be obtained by

\begin{table*}
\setlength{\tabcolsep}{3mm}
\begin{tabular}{l|ccc|ccc|ccc}
\hline
\multirow{2}*{Backbone} & \multicolumn{3}{c|}{STRAT-glasses} &\multicolumn{3}{c|}{STRAT-hat} &\multicolumn{3}{c}{STRAT-tie}\\
& LSSIM↑   & IoU↑  & Disp↓   & LSSIM↑  & IoU↑  & Disp↓ & LSSIM↑  & IoU↑  & Disp↓ \\ \hline \hline
ResNet-18 &  0.6811	& 0.7557    & 0.0150 & 0.5420 & 0.7785 & 0.0195 & 0.2468 & 0.3656 & 0.0147 \\
ResNet-34 &  0.6852	& 0.7571    & 0.0148 & 0.5455 & 0.7859 & 0.0188 & 0.2871 & 0.3893 & 0.0141 \\
ResNet-50 &  \textbf{0.6886}	& \textbf{0.7573} & \textbf{0.0145} & \textbf{0.5470} & \textbf{0.7873}	& \textbf{0.0184} & \textbf{0.2883}	& \textbf{0.3948} & \textbf{0.0131} \\
ResNet-101&  0.6815	& 0.7563	& 0.0148 & 0.5442 & 0.7832 & 0.0191 & 0.2826 & 0.3919 & 0.0140 \\ \hline
\end{tabular}
\caption{The results of our CorrelNet with different backbones on our STRAT dataset. Best results are denoted in boldface.}
\label{tab:backbone}
\end{table*}

\begin{figure*}[t]
    \centering
    \includegraphics[width=1\linewidth]{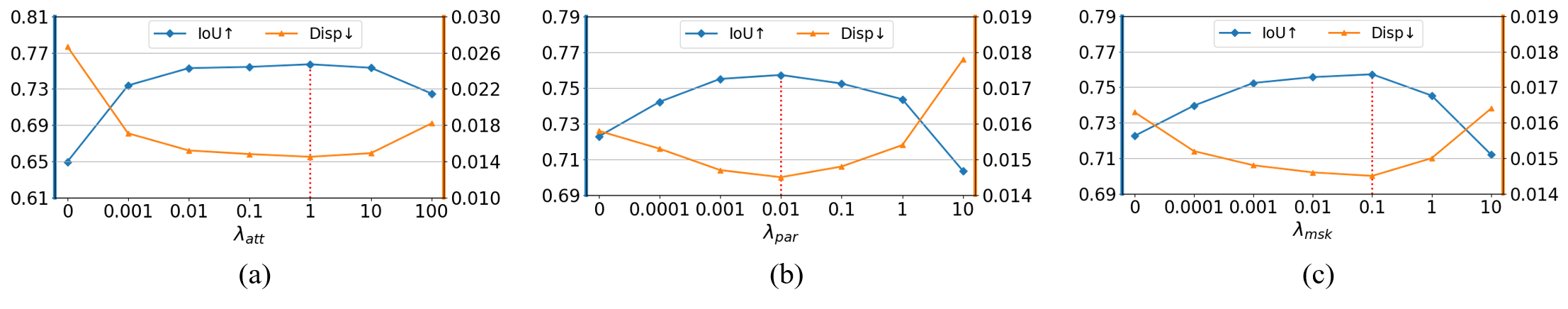}
    \caption{Performance variation of our method with different hyper-parameters $\lambda_{att}$, $\lambda_{par}$, $\lambda_{msk}$ in Eqn. (10) in the main text on our STRAT-glasses dataset. The dashed vertical lines denote the default values used in our other experiments.}\label{fig:hypar_param}
\end{figure*}

\begin{equation}
    \mathbf{I}^c = \tilde{\mathbf{I}}^{f} \odot \tilde{\mathbf{M}}^f + 
    \mathbf{I}^{b} \odot (1 - \tilde{\mathbf{M}}^f),
    \label{eqn:composite}
\end{equation}
where $\odot$ indicates element-wise multiplication. For simplicity, we use $\psi(\mathbf{I}^{f}, \bm{\theta})$ to denote the warped foreground by warping the foreground $\mathbf{I}^{f}$ with parameters $\bm{\theta}$, \emph{i.e.}, $\tilde{\mathbf{I}}^{f}=\psi(\mathbf{I}^{f}, \bm{\theta})$. Then, the notation $\oplus$ is introduced to represent compositing the warped foreground $\tilde{\mathbf{I}}^{f}$ and the background $\mathbf{I}^{b}$. After that, we can rewrite Eqn. (\ref{eqn:composite}) as $\mathbf{I}^c = \psi(\mathbf{I}^{f}, \bm{\theta}) \oplus \mathbf{I}^{b}$, which is exactly Eqn. (1) in the main text.

\section{The Impact of Backbone Network} 
\label{sec:backbone}
In this section, we study the impact of backbone network, the results of which are reported in Table~\ref{tab:backbone}. We change the backbone of our CorrelNet to other ResNet networks \cite{DBLP:conf/cvpr/HeZRS16} with different scales (\emph{e.g.}, ResNet-18, ResNet-34) and evaluate on the three subdatasets of STRAT dataset, respectively. From Table~\ref{tab:backbone}, we find that our method achieves the best result on three subdatasets when using ResNet-50. The performance drop using ResNet-101 might be caused by overfitting. Given the superior performance of ResNet-50, we adopt ResNet-50 as the default backbone.


\section{Hyper-parameter Analyses} \label{sec:hyper_param}
Recall that we have three hyper-parameters in Eqn. (10) of the main text, \emph{i.e.}, cross-attention map loss weight $\lambda_{att}$, warping parameters loss  weight $\lambda_{par}$, and filtering mask loss weight $\lambda_{msk}$, which are respectively set as 1, 0.01, 0.1 via cross-validation by splitting 20\% training samples of STRAT-glasses subdataset as validation set. In this section, we further plot the performance variance of our method on the test set of STRAT-glasses subdataset when varying those hyper-parameters in Figure~\ref{fig:hypar_param}. 

To explore the impact of different cross-attention map loss weight $\lambda_{att}$, we vary $\lambda_{att}$ in the range of [0,100] and the results are shown in Figure~\ref{fig:hypar_param} (a). Comparing the results without cross-attention loss ($\lambda_{att}$=0) and the results with $\lambda_{att}=1$, we can see a clear gap between their performance. When $\lambda_{att}=0$, the quality of predicted cross-attention map cannot be guaranteed without the supervision of ground-truth and low-quality cross-attention map may affect target coordinate prediction, leading to inferior performance. When $\mathcal{L}_{att}$ varies in the range of [0.01, 10], IoU is in the range of [0.7530, 0.7573] and Disp is in the range of [0.0145, 0.0152], which demonstrates that our model is robust when setting $\lambda_{att}$ in a reasonable range. 

With $\lambda_{att}=1$, we experiment with different warping parameter loss weights $\lambda_{par}$ and show the results in Figure~\ref{fig:hypar_param} (b). When $\lambda_{par} \leq 0.01$, the performance increases as $\lambda$ increases, which implies that adding supervision to warping parameters could benefit the spatial transformation performance. When $\lambda_{par}$ becomes larger than 0.01, the performance begins to drop. Moreover, we find that the model can achieve satisfactory results when setting $\lambda_{par}$ in a reasonable range of [0.001, 0.1].

\begin{figure*}
    \centering
    \includegraphics[width=0.95\linewidth]{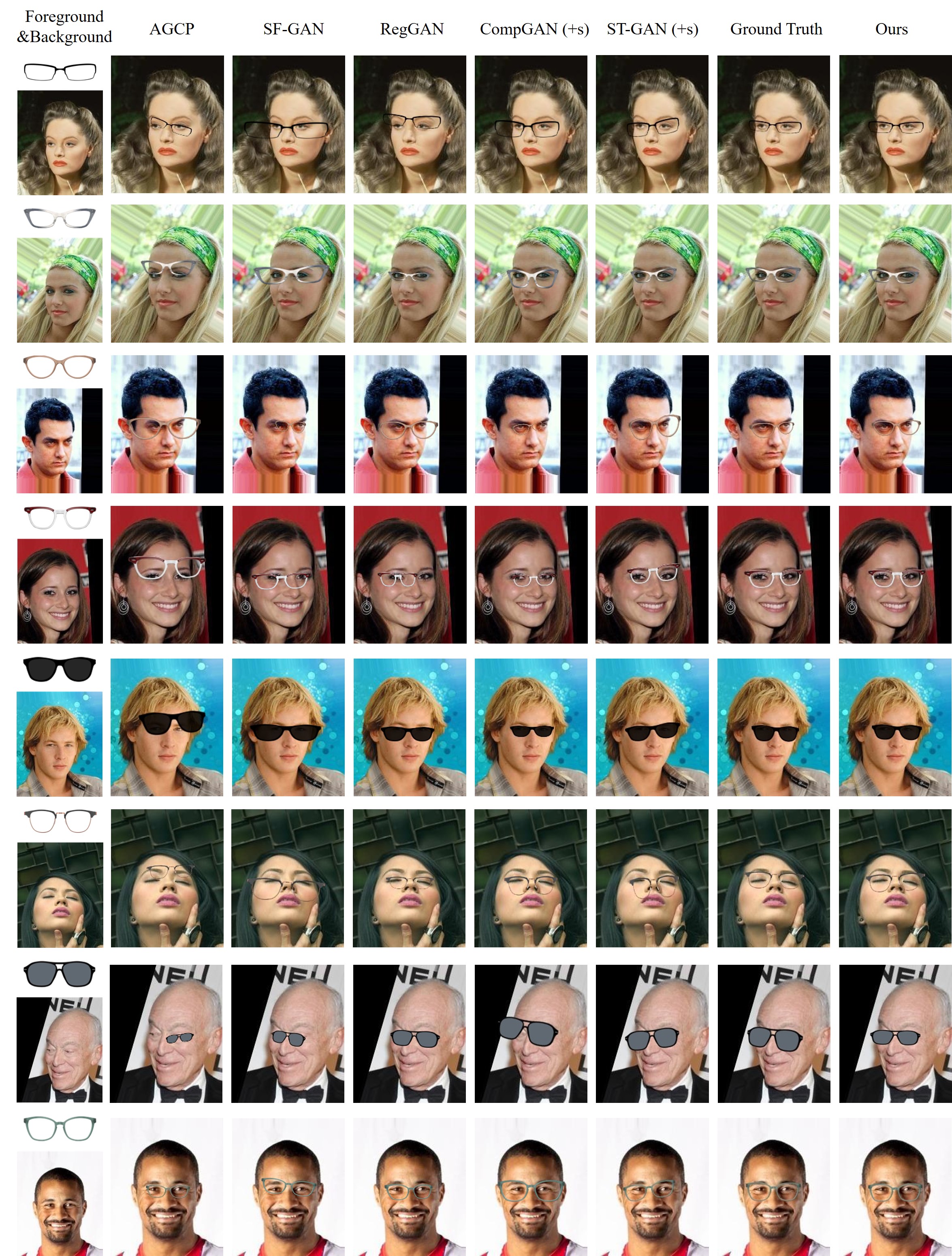}
    \caption{Qualitative comparison of different methods (top row) on STRAT-glasses subdataset. 
    }\label{fig:more_glasses}
\end{figure*}

\begin{figure*}
    \centering
    \includegraphics[width=1\linewidth]{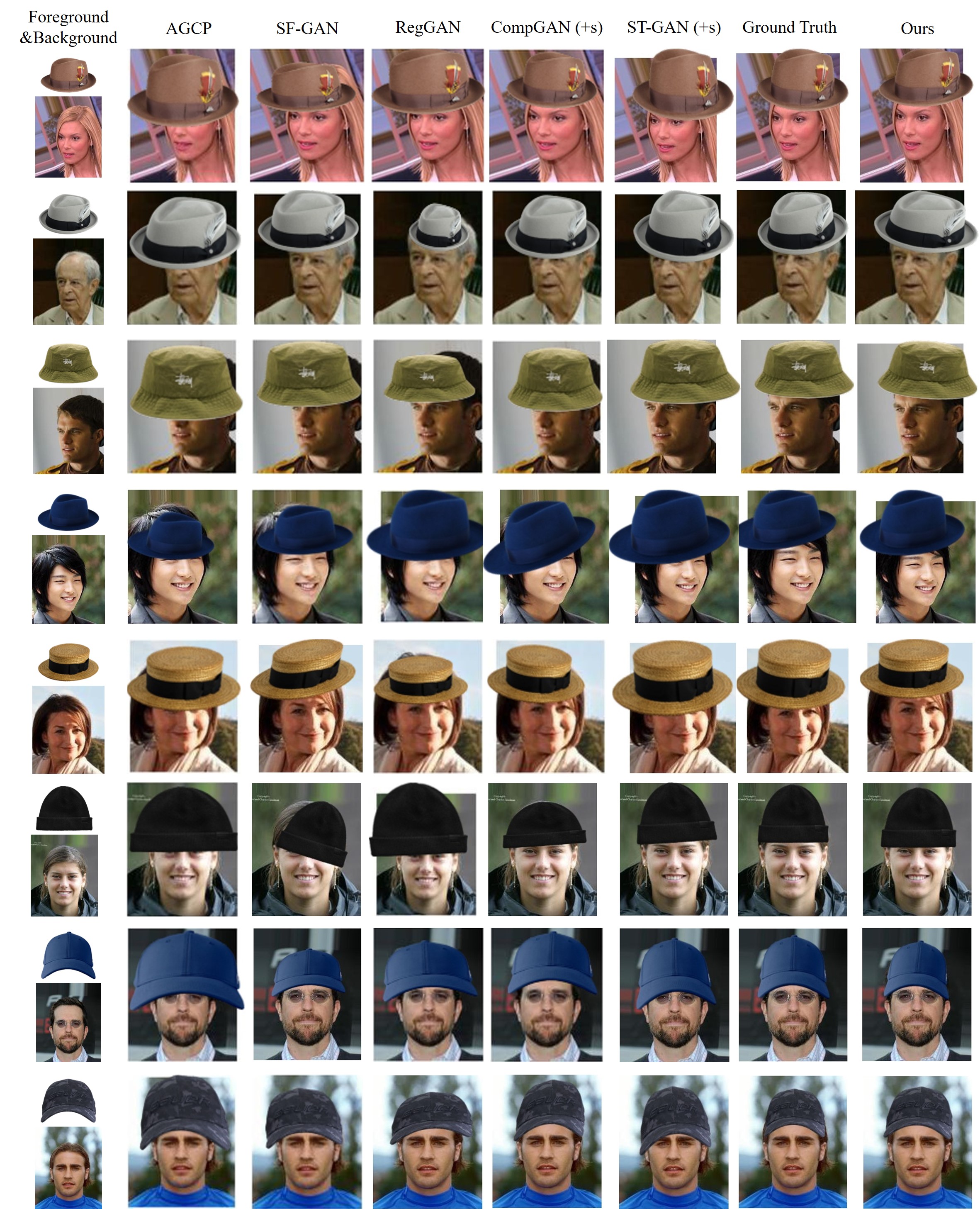}
    \caption{Qualitative comparison of different methods (top row) on STRAT-hat subdataset.
    }\label{fig:more_hat}
\end{figure*}

\begin{figure*}
    \centering
    \includegraphics[width=1\linewidth]{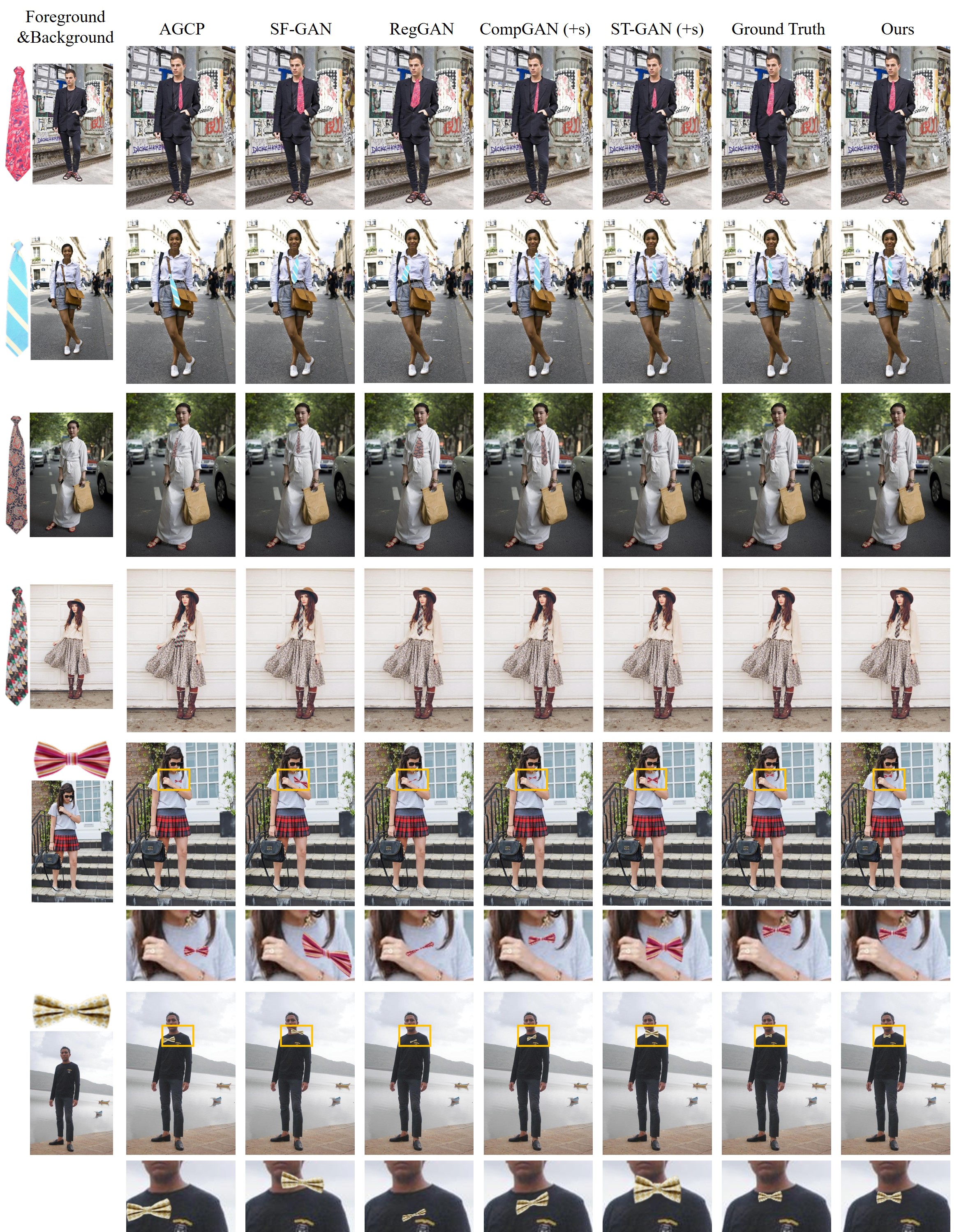}
    \caption{Qualitative comparison of different methods (top row) on STRAT-tie subdataset. In the last two rows, we zoom in for clearer observation.
    }\label{fig:more_tie}
\end{figure*}

By setting $\lambda_{att}=1$ and $\lambda_{par}=0.01$, we further evaluate the performance of our model with different filtering mask loss weights $\lambda_{msk}$, the results of which are shown in Figure~\ref{fig:hypar_param} (c). It can be seen that the model with $\lambda_{msk}=0.1$ clearly outperforms the model with $\lambda_{msk}=0$. Recall we employ the filtering mask loss to force the filtering mask to be negatively correlated with the target prediction error (see Section 5.3 of the main text). When $\lambda_{msk}=0$, the correlation between the predicted filtering mask and the target prediction error map cannot be guaranteed, and the filtering mask may be noisy or even wrong, resulting in a disappointing performance. Moreover, IoU is in the range of [0.7524, 0.7573] and Disp is in the range of [0.0145, 0.0148] when $\lambda_{msk}$ varies in the range of [0.001, 0.1], which implies that our model performs robust to $\lambda_{msk}$ when setting $\lambda_{msk}$ in a reasonable range.    

\section{Visualization of Ablation Studies} 
\label{sec:abltation_study}
In Section 6.4 of the main text, we have evaluated the effectiveness of each component in our method by quantitative comparison. In this section, we qualitatively compare our full-fledged method with different ablated versions: row 1, 2, 3, 5, 8 in Table (2) of the main text, by visualizing some example results produced by those ablated methods on our STRAT dataset in Figure~\ref{fig:ablation_study}.
We can see that the predicted composite images are getting closer to the ground-truth composite images with the addition of proposed loss terms and method designs. 
For example, in column (b) of Figure~\ref{fig:ablation_study}, we only use the warping parameter loss $\mathcal{L}_{par}$ to supervise spatial transformation without calculating cross-attention map, which generally fails to align glasses onto the human face, yielding very poor results. Then, with predicting the target coordinate by using target coordinate loss $\mathcal{L}_{tgt}$, the results in column (c) are far better than column (b), which proves the advantages of predicting target coordinates over predicting warping parameters. Based on column (c), we successively add $\mathcal{L}_{par}$ in column (d), cross-attention map in column (e), from which we can observe gradual improvements that make composite images more plausible. However, the results in column (e) are still far from the ground-truth composite images (column (h)) in terms of foreground shape and size, probably because using all pairs of coordinates to calculate the warping parameter may involve redundant and noisy information. 
Next, based on column (e), by adopting filtering mask to remove noisy pairs of coordinates, our full-fledged method (column (g)) is capable of producing more plausible composite images, which are also closer to the ground-truth.  
Meanwhile, to better demonstrate the utility of the filtering mask, we also show the results that directly use the foreground mask as filtering mask without learning process in column (f), which warps the glasses to unreasonable shapes and the resultant composite images are significantly worse than that of the version without using filtering strategy (column (e)).

In summary, the comparison between ablated methods and full-fledged method proves the effectiveness of predicting target coordinates, capturing foreground-background correspondence, and the proposed filtering strategy.   

\section{More Qualitative Results} 
\label{sec:qualitative_results}

\subsection{Cross-attention Map}
\begin{figure}[t]
    \centering
    \includegraphics[width=0.97\linewidth]{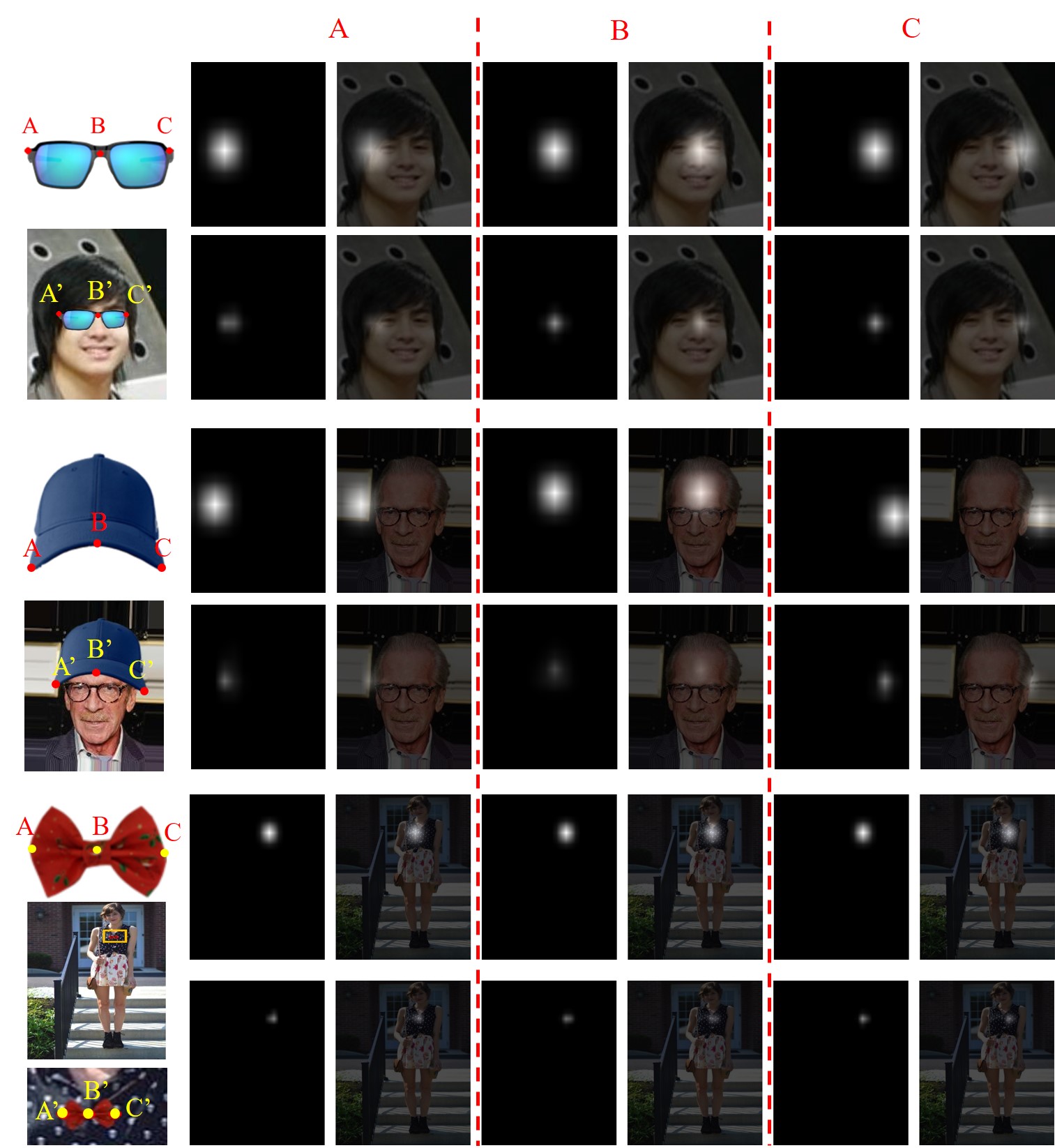}
    \caption{Visualization of the predicted (\emph{resp.}, ground-truth) cross-attention map in the bottom (\emph{resp.}, top) row for three key points (A, B, C). We also superimpose attention map on the background for better visualization. For the composite image in last row, we zoom in for better observation.}\label{fig:cross_attention}
\end{figure}

Recall our method calculates cross-attention map to capture foreground-background correspondence. In Section 6.5.1 of the main text, we visualize the cross-attention map on STRAT-glasses subdataset, in which we select some key points from the foreground image and reshape the corresponding row in the cross-attention map to a 2D map.  
In Figure~\ref{fig:cross_attention}, apart from STRAT-glasses, we additionally provide the cross-attention maps on STRAT-hat and STRAT-tie subdatasets, in which we also mark three keypoints (A,B,C) on each foreground image and show their corresponding row in the cross-attention map.
From Figure~\ref{fig:cross_attention}, we can find that the bright regions in the cross-attention map can roughly match the expected target location on the background. For example, in the first row, the center of the glasses matches the nose on the human face. In the second row, the brim of the baseball cap matches the human forehead. In the last row, it can be seen that the bowtie matches the neck of the person. 
Furthermore, the cross-attention map predicted by our method is close to the ground-truth cross-attention map, which demonstrates that our model is capable of capturing foreground-background correspondence. 

\begin{figure}[t]
    \centering
    \includegraphics[width=0.95\linewidth]{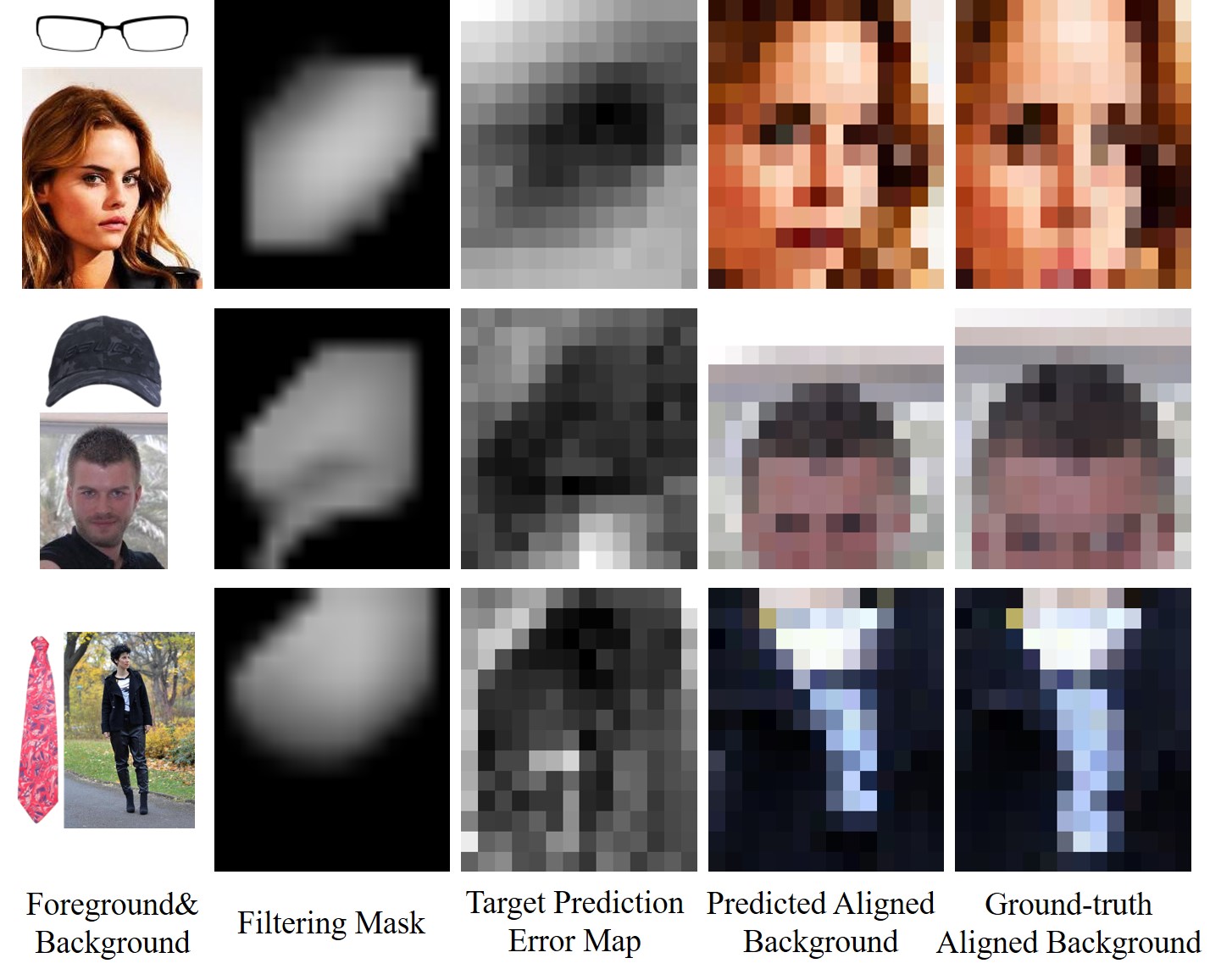}
    \caption{Visualization of predicted filtering mask, target prediction error map, the background images aligned by the predicted/ground-truth target coordinates.
    }\label{fig:filtering_mask}
\end{figure}
\subsection{Filtering Mask}
Recall our method predicts a filtering mask to discard partial noisy pairs of coordinates, in which the filtering mask is expected to be negatively correlated with the target prediction error map (see Section 5.3 of the main text).
Then, in Section 6.5.2 of the main text, we visualize the predicted filtering mask on the STRAT-glasses subdataset. To better demonstrate the effectiveness of the filtering mask, we present additional examples on the three subdatasets of our STRAT dataset in Figure~\ref{fig:filtering_mask}, in which we show foreground image, background image, predicted filtering mask, and target prediction error map that is generated by normalizing the L1 distance between the predicted and ground-truth target coordinates over all locations.    
As shown in Figure~\ref{fig:filtering_mask}, the bright regions in the predicted filtering mask roughly match the dark regions in the target prediction error map, which implies that the predicted filtering mask is negatively correlated with the target prediction error map.
In addition, for intuitive illustration, we also show the aligned background image by assigning the pixel values at predicted/ground-truth target coordinates to the source coordinates in last two columns of Figure~\ref{fig:filtering_mask}.
By comparing the predicted and ground-truth aligned background images, we notice that the regions with more notable misalignment generally match the dark regions in the filtering mask.
For example, from the predicted and ground-truth aligned background images in the first row of Figure~\ref{fig:filtering_mask}, we see that the surrounding regions on the human face (\emph{e.g.}, hair and mouth) are obviously misaligned, which roughly match the dark regions in the filtering mask.
In the last row of Figure~\ref{fig:filtering_mask}, there is noticeable misalignment in the lower regions of the predicted and ground-truth aligned background images, which also corresponds with the lower dark regions in the filtering mask.   
These examples further shed light on the validity of the proposed filtering strategy, which is able to successfully discard the noisy pairs of coordinates, achieving more accurate warping parameters estimation.

\begin{figure}[t]
    \centering
    \includegraphics[width=1\linewidth]{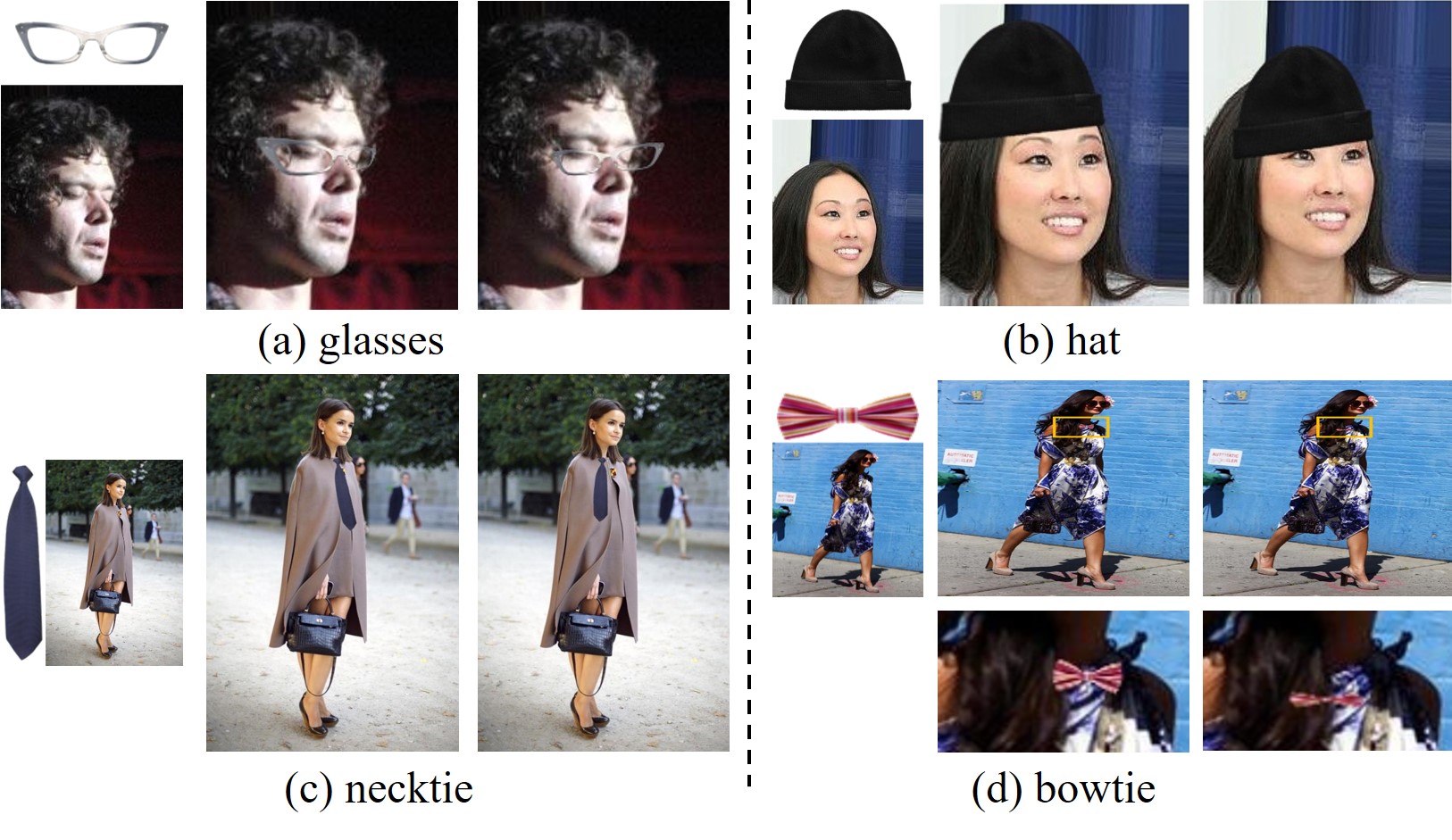}
    \caption{Visualization of failure cases produced by our method on STRAT dataset. In each example, from left to right: foreground and background images, ground-truth composite image, predicted composite image. For bowtie (d), we zoom in for clearer observation.}
    \label{fig:failure_cases}
\end{figure}

\subsection{Composite Results}
As described in Section 3 of the main text, our STRAT dataset consists of three subdatasets with each subdataset containing total of 3000 pairs of foregrounds and backgrounds.
When constructing STRAT dataset, we collect 30 different foreground objects for each subdataset and ensure the diversity of the persons on the background in terms of age, gender, skin color, face orientation, posture, clothing style, \emph{et al}. 
Partial examples of STRAT dataset are shown in Figure~\ref{fig:more_glasses}, Figure~\ref{fig:more_hat}, Figure~\ref{fig:more_tie}, which demonstrate that our STRAT dataset is diverse enough to cope with various real-world scenarios.

Next, we show more resultant composite images of our CorrelNet and five baseline methods on three subdatasets, \emph{i.e.}, STRAT-glasses, STRAT-hat, STRAT-tie, in Figure~\ref{fig:more_glasses}, Figure~\ref{fig:more_hat}, Figure~\ref{fig:more_tie}, respectively. These baseline methods include AGCP \cite{Li2021ImageSV}, SF-GAN \cite{SF-GAN}, RegGAN \cite{RegGAN2019}, CompGAN (+s) \cite{Compositional-GAN}, and ST-GAN (+s) \cite{stgan}. ST-GAN (+s) (\emph{resp.}, CompGAN (+s)) is the enhanced version of ST-GAN \cite{stgan} (\emph{resp.}, CompGAN \cite{Compositional-GAN}), in which we add direct supervision on the composite image.     
For given various foreground and background images, it can be seen that our method can typically produce realistic and plausible composite images that are close to the ground-truth composite images, while other baseline methods may warp foreground objects to unreasonable positions or shapes. 
Take the fourth row of Figure~\ref{fig:more_glasses} as example, most baseline methods fail to align the glasses to the eyes on the human face, while our method works well, generating more realistic composite image.
In the third-to-last row of Figure~\ref{fig:more_hat}, most compared methods warp the hat to unreasonable shapes or positions, leading to unconvincing composite images. In contrast, the proposed method roughly preserves the shape of the hat during warping foreground, which makes the resultant composite image more plausible.
Moreover, as shown in the last row of Figure~\ref{fig:more_tie}, it is supposed to align the bowtie to the neckline of the person on the background. However, most methods fail on it, but our method accomplishes it correctly.
Those qualitative comparisons further confirm the superiority of our proposed method on spatial transformation for image composition.       

\section{Discussion on Limitation} 
\label{sec:limitation}
Although our method can generally achieve satisfactory results, it may fail on some challenging cases. For example, as shown in Figure~\ref{fig:failure_cases} (a), our model fails to generate plausible composite image, in which the warped foreground does not match the human face in the background. This is probably because that the misaligned face and closed eyes make wearing glasses properly more challenging. 
In Figure~\ref{fig:failure_cases} (b), our method tends to stick the hat to the forehead of the human in the background, resulting in an unrealistic composite image, which may be due to the special shape of the hat and the non-frontal orientation of the human face. 
Additionally, in Figure~\ref{fig:failure_cases} (c) and (d), the foreground objects are warped to unreasonable positions or shapes, which may be attributed to the non-frontal posture of the human in background that inevitably increases the difficulty of wearing necktie or bowtie correctly.

{\small
\bibliographystyle{ieee_fullname}
\bibliography{egbib}
}